\documentclass{article}

\usepackage{arxiv}

\usepackage[utf8]{inputenc} 
\usepackage[T1]{fontenc}    
\usepackage{hyperref}       
\usepackage{xr-hyper}
\usepackage{url}            
\usepackage{cite}
\usepackage{booktabs}       
\usepackage{tabularx}       
\usepackage{amsfonts}       
\usepackage{nicefrac}       
\usepackage{microtype}      
\usepackage{lipsum}
\usepackage{graphicx}
\usepackage{tabularx}
\graphicspath{ {./images/} }
\usepackage{array}
\usepackage{float}
\usepackage{chemformula} 
\usepackage{chemmacros} 

\makeatletter

\newcommand{\Rmnum}[1]{\expandafter\@slowromancap\romannumeral #1@}
\makeatother

\title{Emission-GPT: A domain-specific language model agent for knowledge retrieval, emission inventory and data analysis}

\author{
\textbf{Jiashu Ye\textsuperscript{1}, Tong Wu\textsuperscript{1}, Weiwen Chen\textsuperscript{1}, Hao Zhang\textsuperscript{1}, Zeteng Lin\textsuperscript{3}, Xingxing Li\textsuperscript{3}} \\
\textbf{Shujuan Weng\textsuperscript{1}, Manni Zhu\textsuperscript{1},  Xin Yuan\textsuperscript{2}, Xinlong Hong\textsuperscript{2}, Jingjie Li\textsuperscript{2},}\\
\textbf{Junyu Zheng\textsuperscript{1,*}, Zhijiong Huang\textsuperscript{2,*},  Jing Tang\textsuperscript{3,*}} \\
\\
\textsuperscript{1}Sustainable Energy and Environment Thrust, Function Hub\\
The Hong Kong University of Science and Technology (Guangzhou)\\
Guangzhou 511400, China \\
\\
\textsuperscript{2}College of Environment and Climate\\
Institute for Environment and Climate Research\\
Jinan University\\
Guangzhou 511443, China\\
\\
\textsuperscript{3}Data Science and Analytics Thrust, Information Hub\\
The Hong Kong University of Science and Technology (Guangzhou)\\
Guangzhou 511400, China \\
\\
\textsuperscript{*}Corresponding author: \texttt{junyuzheng@hkust-gz.edu.cn}, \texttt{huangzj@jnu.edu.cn}, \texttt{jingtang@hkust-gz.edu.cn}
}

\begin{document}
\maketitle
\begin{abstract}
Improving air quality and addressing climate change relies on accurate understanding and analysis of air pollutant and greenhouse gas emissions. However, emission-related knowledge is often fragmented and highly specialized, while existing methods for accessing and compiling emissions data remain inefficient. These issues hinder the ability of non-experts to interpret emissions information, posing challenges to research and management. To address this, we present Emission-GPT, a knowledge-enhanced large language model agent tailored for the atmospheric emissions domain. Built on a curated knowledge base of over 10,000 documents (including standards, reports, guidebooks, and peer-reviewed literature), Emission-GPT integrates prompt engineering and question completion to support accurate domain-specific question answering. Emission-GPT also enables users to interactively analyze emissions data via natural language, such as querying and visualizing inventories, analyzing source contributions, and recommending emission factors for user-defined scenarios. A case study in Guangdong Province demonstrates that Emission-GPT can extract key insights\textemdash such as point source distributions and sectoral trends\textemdash directly from raw data with simple prompts. Its modular and extensible architecture facilitates automation of traditionally manual workflows, positioning Emission-GPT as a foundational tool for next-generation emission inventory development and scenario-based assessment.

\end{abstract}
\keywords{Emission \and Large language model \and Toolchain agent}

\newpage

{\LARGE \textbf{Graphical Abstract}}

\vspace{1em}

\begin{figure}[H] 
    \centering
    \includegraphics[width=1\textwidth]{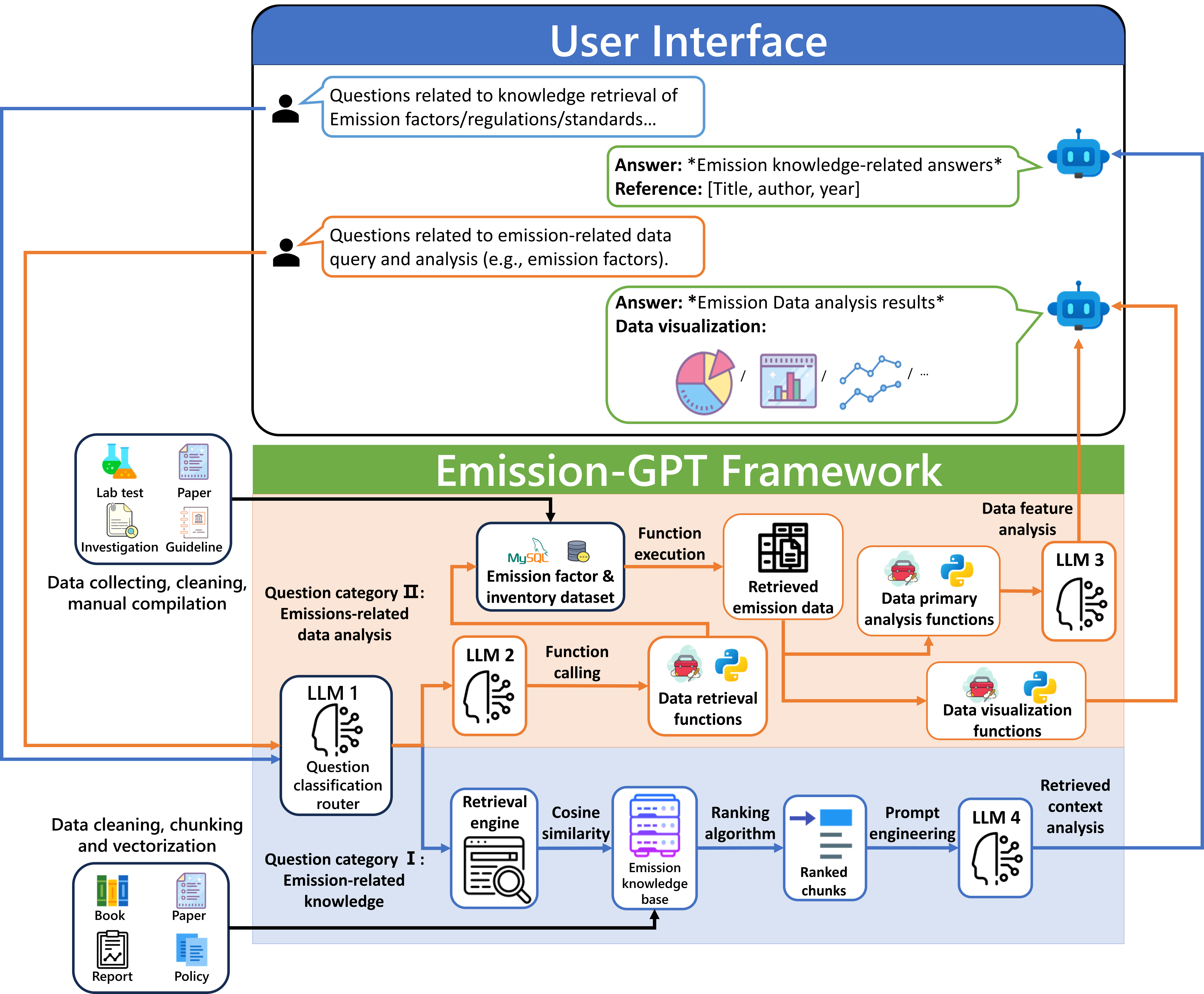}
\end{figure}

\newpage

\section{Introduction}


Anthropogenic emissions, a complex mixture of greenhouse gases, reactive trace gases, and aerosols released into the atmosphere, are the direct drivers of climate change and pervasive air pollution \cite{arshad2024air,WOS:001270419200003,WOS:000171538500007}. A comprehensive understanding of emissions, spanning their underlying physical and chemical properties, governing standards, emission quantities, and formation mechanisms, is therefore fundamental to both scientific progress and effective environmental policy \cite{WOS:000362603400040,WOS:001458965700001}. In research, this knowledge provides the essential context for interpreting atmospheric observations and developing robust predictive models, such as Chemical Transport Models (CTMs) and Earth System Models (ESMs), for which emission inventories serve as a critical input \cite{WOS:001059168800003,WOS:000444918100034,WOS:000863776000001,WOS:000989940900001,WOS:000515601700001}. In the realm of environmental management, a deep understanding of emission sources, control technologies, and regulatory limits is indispensable for formulating targeted abatement strategies, performing cost-benefit analyses, and evaluating policy efficiency \cite{WOS:000659774800001,WOS:001391097700001}.

Despite its critical importance, acquiring and synthesizing emissions information is a fragmented, labor-intensive, and inefficient process. This challenge extends far beyond the compilation of emission inventories to include fundamental tasks such as identifying appropriate emission factors and emission estimation methodologies, interpreting technical guidelines, and understanding regulatory standards \cite{WOS:001419642600001,WOS:000794098000001,WOS:000678317000001,WOS:000720706300002,WOS:001139176700001}. Professionals, from researchers to government staff, must navigate a complex web of varied sources, including dense technical reports, evolving government regulations, and vast academic literature, to find the necessary and appropriate data and context \cite{WOS:000171538500007,WOS:000279743500009,WOS:000373808000002}. This effort demands a substantial time and deep domain expertise, such as an intricate knowledge of industrial processes, source classification systems, and abatement technologies \cite{WOS:000720706300002,WOS:001050201300001,WOS:000592370700005,WOS:001496463300001,WOS:000663572800003,WOS:001271680800001}.

The analysis and interpretation of this complex emission information also present challenges, particularly for stakeholders who are not domain specialists \cite{WOS:001020542300001,WOS:000477934800022,WOS:000740208500011,WOS:000170964900006}. Researchers may struggle to align sector-specific emissions data with applicable standards \cite{RN3}; policymakers may lack rapid access to synthesized regional trends \cite{RN2}; and the public is often overwhelmed by technical jargon and fragmented information \cite{RN1}; As a result, the ability to extract actionable insights from emission data remains concentrated among a small number of experts, limiting its broader application in science, governance, and public engagement.

Recent advances in artificial intelligence, particularly large language models (LLMs), have enabled powerful capabilities in natural language processing, offering new solutions for complex challenges in information retrieval and integration \cite{10729345,10925029,9076502,10581180}. To address persistent difficulties in managing atmospheric emission knowledge, we developed \textbf{Emission-GPT}, a domain-specific agent that combines LLMs with retrieval-augmented generation (RAG), function calling, and few-shot chain-of-thought (CoT) reasoning \cite{10930066,10779705}. General-purpose LLMs often hallucinate and lack precision in technical domains due to static training data and limited domain knowledge \cite{10845736,10781509,10963259,10791375}. Emission-GPT mitigates these issues by leveraging a curated knowledge base of over 10,000 documents\textemdash including peer-reviewed articles, technical reports, standards, and reference books\textemdash segmented into over one million 256-token chunks for efficient retrieval. It delivers three core functions: (1) accurate responses to queries on inventory methodologies, pollutants, and policies; (2) context-aware emission factor recommendations with source traceability; and (3) natural language analysis of emission inventory datasets for intuitive, multi-dimensional exploration. 

This paper details the motivation, design, and initial implementation of the Emission-GPT system. We demonstrate its potential to lower the technical barriers associated with emission-related work and enhance the overall value derived from such data. The remainder of this paper is structured as follows: Section \ref{Data and Methods} describes the system design of Emission-GPT. Section \ref{Results and discussion} elaborates on the performance evaluation and showcases and concludes with a discussion of implications and future directions.

\section{Data and methods}
\label{Data and Methods}

\subsection{Pipeline for Emission-GPT}

\begin{figure} 
    \centering
    \includegraphics[width=0.8\textwidth]{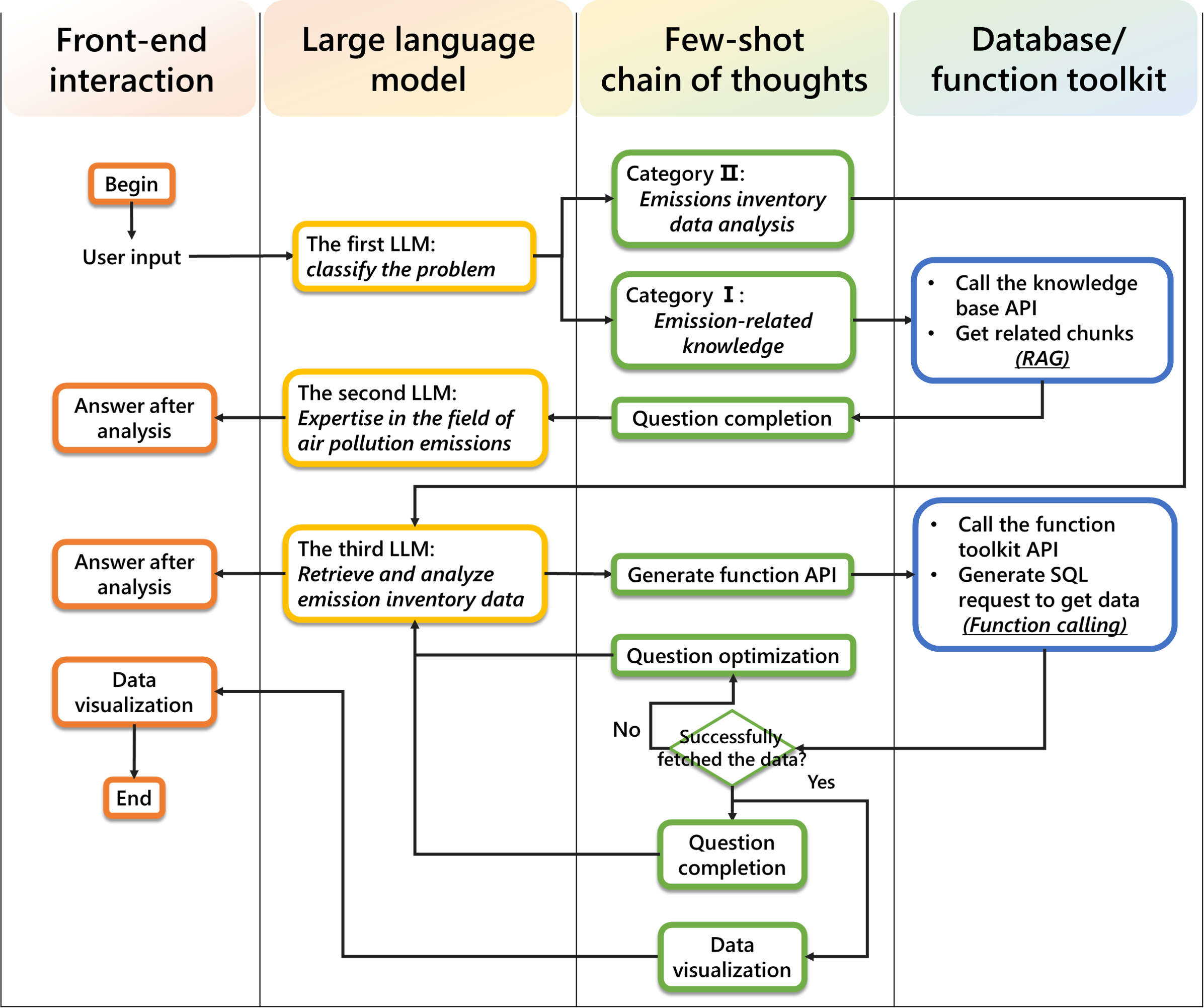}
    \caption{Pipeline for Emission-GPT.}\label{fig:pipeline}
\end{figure}

This study proposes a modular, multi-stage workflow designed to enable intelligent interaction and analysis in the domain of atmospheric emissions, leveraging LLMs in combination with retrieval and function-calling capabilities. As Figure \ref{fig:pipeline} shows, the framework is structured into four major components: Front-end interaction, large language model module, intermediate processing, and database/function toolkit. Upon receiving a user query, the first-stage LLM classifies the question into one of two categories: (\Rmnum{1}) emission-related knowledge or (\Rmnum{2}) emission related data analysis.

For Category \Rmnum{1} queries, a second-stage expert LLM performs knowledge retrieval using RAG, accessing a structured knowledge base to supplement the model's response and complete the answer. In contrast, for Category \Rmnum{2} queries, a third-stage LLM is invoked to construct API-level requests and SQL-like queries that interface with backend emission inventory and emission factor databases via function calling. These queries are subject to optimization if data retrieval fails, ensuring robustness. Upon successful data acquisition, the system performs analysis and optionally visualizes the results for the end-user. This modularized pipeline integrates domain-specific expertise, automated reasoning, and data-driven analysis to support both qualitative and quantitative insights in the field of atmospheric emission. Moreover, its flexible design allows for future extension to additional categories of emission-related tasks, each supported by corresponding workflows or specialized AI agents.

\subsection{Dataset preparation and preprocessing} 

To support Emission-GPT’s knowledge-intensive tasks, we constructed a domain-specific knowledge base comprising 10,332 authoritative documents related to atmospheric emissions. The corpus includes 9,493 peer-reviewed journal articles, 809 policy and technical documents, and 30 scholarly books in both Chinese and English. To ensure the quality, authority, and representativeness of the corpus, the data collection involved extensive manual effort. A total of 24 doctoral and master’s students participated in a month-long curation task, manually filtering and organizing documents from multiple authoritative sources, including the Web of Science Core Collection, China National Knowledge Infrastructure (CNKI), the United States Environmental Protection Agency (EPA), the Ministry of Ecology and Environment (MEE) of China, and the Intergovernmental Panel on Climate Change (IPCC). Each participant was responsible for specific time periods, source types, and topical categories. This large-scale, labor-intensive process ensured relevance, consistency, and traceability across all entries.

The curated knowledge base is systematically structured by document type, region, emission source, pollutant, and research theme. Covers major sectors (industrial, agricultural, biomass burning), key pollutants (\(\mathrm{CO_2}\), \(\mathrm{NO_x}\), \(\mathrm{PM_{2.5}}\)), and various geographic scales. Reports span from 1972 to 2025, while books provide foundational references on inventory methods, atmospheric chemistry, and modeling. For a detailed breakdown, see Knowledge Base Composition and Topical Coverage section.

A customized multi-stage preprocessing pipeline was implemented to handle document heterogeneity. Layout detection was optimized for scientific structures, including multi-column formats, inline mathematical notations, and nested tables. Scanned materials (23\%) were processed using an optical character recognition (OCR) workflow integrating Paddle Optical Character Recognition (Paddle-OCR) with region-specific masking, achieving 94.7\% character-level accuracy. Technical content was further refined using Unified Multimodal Encoder-Decoder for Mathematical Expression Recognition Network (UniMERNet) for formulas and a hybrid Structured Equation Table (StructEqTable) method for emission-related tables. Postprocessing included intelligent merging of split references, filtering of policy boilerplate, and normalization of emission-specific entities (e.g., maintaining subscripts in $CO_2$, $CH_4$). The final corpus comprises 4.7 million structured text segments with metadata annotations, achieving 98.2\% content fidelity, and serves as the knowledge base for the RAG pipeline of Emission-GPT.

\subsection{Workflow of RAG}
In the RAG framework, a user query is first transformed into a vector and used to retrieve relevant information from the previous established knowledge base \cite{NEURIPS2020_6b493230,wang2024bioragragllmframeworkbiological}. In this study, we use the natural language understanding model Qwen-plus for context segmentation and the pretrained embedding model BGE-M3 to generate dense vector representations of both queries and document chunks \cite{qwen,chen2024bgem3embeddingmultilingualmultifunctionality}. Document passages are ranked based on cosine similarity between embeddings, allowing semantically relevant content to be retrieved more accurately than traditional keyword-based approaches, particularly for ambiguous or information-dense queries \cite{RN113,6705226}.

The generator model Qwen-plus then synthesizes responses by combining the original query with the retrieved passages \cite{qwen}. By grounding outputs in retrieved external knowledge, this approach enhances factual accuracy, contextual relevance, and language fluency. Compared to standard generative models, the RAG generator achieves higher answer consistency and credibility through evidence-based generation \cite{jiang-etal-2023-active}.

RAG facilitates multi-turn conversations by embedding previous queries and responses into the context of each new interaction \cite{li2025lexragbenchmarkingretrievalaugmentedgeneration}. This iterative framework allows the system to continuously refine its retrieval and generation processes in response to user input \cite{cheng2024coralbenchmarkingmultiturnconversational}. Over time, the feedback loop enhances alignment with user intent, making it especially effective for complex tasks that involve multi-step reasoning or require the integration of information across multiple turns \cite{tan2024peerreviewmultiturnlongcontext}.

\subsection{Emission factors recommendation}

Emission factors (EFs) serve as essential foundational data in emission inventory compilation, directly influencing the accuracy of pollutant and greenhouse gas emission estimates. However, traditional approaches to selecting emission factors face significant challenges, such as being highly time-consuming, demanding extensive professional expertise, and presenting substantial barriers to entry, all of which can negatively impact inventory accuracy. To improve efficiency, reduce complexity, and lower barriers to effective EF selection, we developed an emission factor recommendation tool leveraging generative artificial intelligence. We demonstrate the utility of this tool through a case study focused on mobile source emissions, specifically vehicle exhaust emissions. The following sections detail the methodological development and key features of the proposed recommendation tool.

\subsubsection{Providing targeted references to standards and guidelines}

To ensure the credibility, transparency, and traceability of the recommended emission factors (EFs), the system adopts a structured two-stage retrieval and evaluation framework \cite{usepa2025}.

In the first stage, the system performs schema-based matching between the user's specified source attributes and officially published technical guidelines or regulatory standards. Emission factors retrieved from these authoritative documents are included in the recommendation list without further evaluation, serving as high-trust, policy-aligned references. In the second stage, to address cases where official values are unavailable, incomplete, or insufficiently specific, the system conducts a semantic retrieval across curated peer-reviewed literature and publicly available EF databases. Retrieved records are then subjected to a standardized ranking procedure, which evaluates candidates based on predefined criteria such as methodological robustness, data representativeness, and source credibility. The final output includes the top-ranked EF entries, each accompanied by pollutant values, metadata on testing methods, and bibliographic provenance.

This dual-stage approach enables the system to prioritize regulatory compliance while also incorporating empirically rich, context-relevant factors, thereby supporting scientifically grounded and operationally applicable EF selection.

\subsubsection{Querying for emission factors}

In this study, an LLM tool was introduced to enable interactive and condition-driven retrieval of emission factors. When a user initiates a query, the LLM first engages in natural language interaction to guide the user in specifying key attributes of the emission source, including vehicle type, fuel type, emission standard, and geographic region. This guided input process ensures that the retrieved information is structurally complete and semantically standardized.

Once all necessary parameters are collected, the LLM uses the provided information to retrieve candidate emission factors from a systematically compiled emission factor database, constructed from both domestic and international authoritative sources, including the Technical Guide for Compiling Integrated Air Pollutant and Greenhouse Gas Emission Inventories (Trial), Technical Guideline on City-level Air Pollutant Emission Inventory Development, the IPCC Guidelines for National Greenhouse Gas Inventories, as well as peer-reviewed literature \cite{MEE2024fusion,MEE2024city,ipcc2019}. Leveraging its semantic understanding and contextual reasoning capabilities, the model performs meaning-level matching to identify the most relevant EF entries corresponding to the user input.

\subsubsection{Ranking the recommended emission factors}

To ensure that the recommended emission factors (EFs) are both scientifically robust and contextually appropriate, a multi-criteria evaluation and ranking framework was established. This framework enables transparent comparison among candidate EFs and enhances the reliability of the final recommendations.

The evaluation system comprises four key dimensions: data representativeness, methodological reliability, sample representativeness, and data authority \cite{ciroth2021data}. Each EF is qualitatively assessed on a four-level grading scale (Grades A to D), with Grade A indicating the highest level of data quality. The detailed grading criteria (outlining specific characteristics of each grade across all dimensions) are provided in Table \ref{table: EF_recommendation}. These criteria were developed based on a comprehensive literature review, expert consultation, and reference to established EF data quality standards \cite{ntziachristos2017emep}.

To quantify the overall quality of each emission factor, relative weights were assigned to the four evaluation dimensions according to their contribution to EF reliability and applicability. Specifically, data representativeness and methodological reliability were each weighted at 35\%, sample representativeness at 20\%, and data authority at 10\%. These weights were determined through consultation with domain experts and internal testing to ensure robust and balanced evaluation outcomes \cite{EPA2024webfire}. Each qualitative grade was then mapped to a numerical score (A = 4, B = 3, C = 2, D = 1). The final composite score for each EF is calculated as follows:

\begin{equation}\label{eq: EF score}
    \sum^n_{i=1} w_i \times s_i
\end{equation}

Where, $s_i$ represents the score of the $i$ dimension, and $w_i$ is the corresponding weight. Based on the final scores, all emission factors were ranked in descending order. The top five scoring factors were selected as the recommended emission factors. These factors represent the most reliable and contextually appropriate options according to the evaluation criteria.

\subsection{Tools design and prompt engineering}

Traditional emission analysis tools require specialized knowledge and coding skills, limiting accessibility. To address this, Emission-GPT integrates modular tools and prompt engineering, enabling users to query and analyze emission data through natural language, while ensuring reproducibility, and operational efficiency.

\subsubsection{Tool functionality design}

To enable domain-specific operations within the Emission-GPT system, we designed a suite of modular functions with standardized interfaces and descriptive metadata. Each function is defined at initialization with a unique name, a concise description of its purpose, and a formal input schema based on the JSON format. During runtime, the language model dynamically selects the appropriate tool by generating a structured JSON object that specifies the function name and its required parameters.

A CoT layer interprets the function call, executes the corresponding backend procedure, and returns the output\textendash such as tabular data, statistical summaries, or interactive visualizations (Figure \ref{fig:pipeline}). These routines interface directly with emission related databases, enabling pollutant-specific data retrieval, and execution of analytical operations. Results are rendered in user-friendly formats, including structured tables and charts.

This design enables users to retrieve emission-related data and perform analytical tasks using natural language instructions, significantly lowering the technical barrier to data access. By coupling language-based reasoning with structured backend functions, the system streamlines analytical workflows, enhances operational efficiency, and ensures the reproducibility of each query-response interaction.

\subsubsection{Prompt engineering for structured invocation}

To ensure accurate and trustworthy responses, the system dynamically reconstructs prompt contexts for each user interaction by retrieving semantically relevant segments from a curated knowledge base. These segments are clearly marked to ground the model’s output in verifiable sources. If no adequate context is found, a fallback instruction prompts the model to return an explicit notice rather than a speculative answer.

\begin{figure}[ht] 
    \centering
    \includegraphics[width=0.6\textwidth]{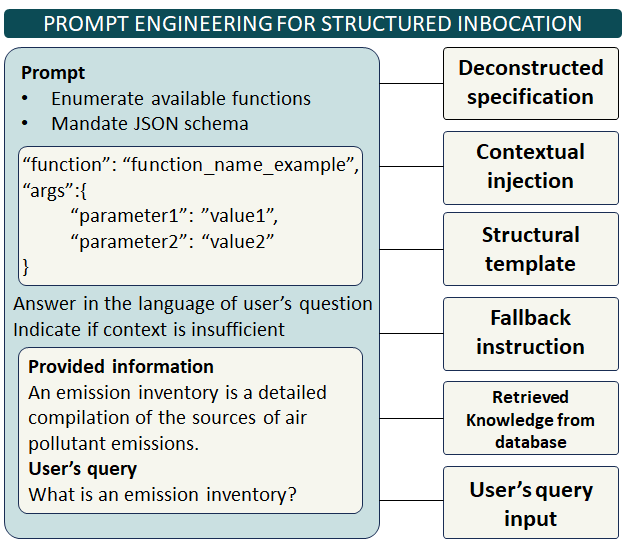}
    \caption{Structured prompt engineering framework for controlled function invocation. }\label{fig:prompt_structure}
\end{figure}

Following generation, all outputs are validated against predefined function metadata\textemdash including function names, schemas, and parameter formats\textemdash to ensure structural compliance (Figure~\ref{fig:prompt_structure}). Invalid outputs trigger clarification protocols rather than silent failure. This integrated framework establishes a reliable bridge between natural language queries and deterministic computational operations, enabling consistent, transparent, and reproducible support for complex environmental tasks.

\section{Results and discussion}
\label{Results and discussion}

\subsection{Model performance evaluation and benchmark datasets}

To systematically evaluate the performance of Emission-GPT, we adopted a two evaluation strategy combining both objective metrics and expert-driven subjective analysis.

To objectively evaluate the RAG pipeline, we developed a benchmark dataset focused on air-emission-related tasks and applied the Retrieval-Augmented Generation Assessment (RAGAS) framework within EvalScope. The dataset comprises 60 expert-curated questions across three difficulty levels (Level 1\textendash 3, representing increasing complexity) and four thematic categories: (1) emission concepts and definitions, (2) emission standards, (3) measurement techniques, and (4) emission inventories and data analysis. Each question is paired with a reference answer and supporting context to enable detailed evaluation. Generated responses were assessed using six dimensions (faithfulness, answer relevance, semantic similarity, context relevance, context precision, and context recall) providing a comprehensive evaluation of factual accuracy, contextual alignment, and semantic consistency. 

In parallel, a semi-quantitative human evaluation was conducted by multiple domain experts. A total of 60 domain-relevant evaluation questions were initially designed, covering realistic, domain-specific scenarios, and answers were generated by different language models. From the resulting question-answer pairs, 15 were randomly selected across groups to conduct a comparative evaluation. Three representative models\textemdash Emission-GPT, GPT-4o, and DeepSeek R1\textemdash were independently tested. These questions were carefully developed to mirror realistic, domain-specific scenarios commonly encountered in practice, encompassing four thematic areas: emission inventory compilation, retrieval of technical standards, source-level testing, and integration of emissions-related knowledge. The questions were further categorized according to three difficulty levels.

Responses from each model were independently reviewed by a panel of 8 PhD-level experts specializing in environmental engineering and air quality management. The evaluation specifically emphasized factual accuracy and relevance within the domain context, providing a critical, human-centered complement to automated performance metrics.

\subsection{Automatic evaluation of knowledge base performance}

\begin{figure} 
    \centering
    \includegraphics[width=0.7\textwidth]{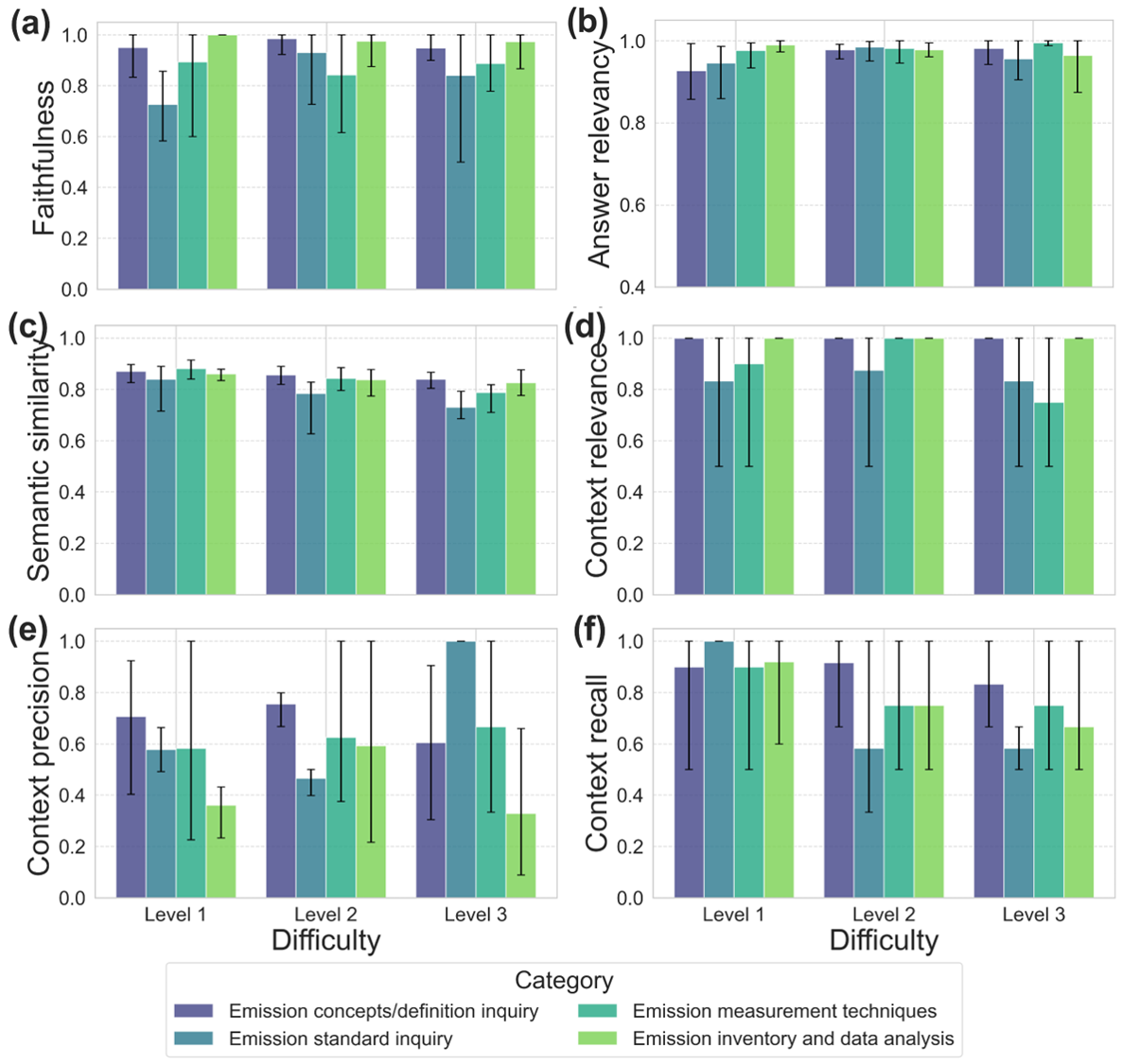}
    \caption{Model performance automatic evaluation of Emission-GPT. (a–-f) EvalScope evaluation metrics (average faithfulness, answer relevancy, semantic similarity, context relevance, context precision, context recall) of Emission-GPT under different question categories and question difficulties.}\label{fig:EvalScopeMetrics}
\end{figure}

Building on a richly structured knowledge base comprising over 10,000 documents\textemdash including journal articles, institutional reports, and academic books\textemdash Emission-GPT benefits from broad topical coverage across emission sectors, pollutants, and geographic regions (see Text \ref{Knowledge Base Composition and Topical Coverage}). To evaluate the effectiveness of this knowledge base, we conducted an automatic performance assessment using six quantitative metrics. The quantitative evaluation results are shown in Figure \ref{fig:EvalScopeMetrics}. Overall, generation-level metrics such as faithfulness, answer relevancy, semantic similarity, and context relevance exhibit consistently high scores across most categories and difficulty levels (Figure \ref{fig:EvalScopeMetrics}a-d).

Specifically, average answer relevancy remains above 0.90 with minimal variance, indicating that the generated answers are well-aligned with user queries regardless of domain or complexity \cite{es2025ragasautomatedevaluationretrieval}. Similarly, faithfulness, semantic similarity, and context relevance scores typically exceed 0.70. However, greater variability is observed in the categories of "Emission Measurement Techniques" and “Emission Standard Inquiry”, which is likely attributed to the inclusion of excessive retrieved context. This may have introduced a substantial amount of irrelevant or weakly related content into the evaluation, thereby increasing score dispersion \cite{es2025ragasautomatedevaluationretrieval,knollmeyer2024benchmarking,purwar2024evaluating}.

In contrast, context precision scores show greater variability and lower averages across categories and difficulty levels, ranging from approximately 0.15 to 1.00 (Figure \ref{fig:EvalScopeMetrics}e). Context precision measures the proportion of relevant context chunks among all retrieved chunks. Low scores indicate that retrieved contexts contain limited relevant information \cite{b2024evaluatingefficacyopensourcellms,Morgan_2024}. We found that mixed Chinese-English content exist in retrieved chunks, which might significantly reduces context precision accuracy. This is largely due to the limitations of current evaluation frameworks in handling multilingual contexts, which impairs their ability to accurately assess precision.

In summary, the language model performs well in generating accurate and relevant responses when provided with appropriate context. However, its overall robustness is limited by retrieval deficiencies, particularly for complex or information-rich queries. To address this, future work should aim to enhance retriever precision and recall through advanced semantic search methods, adaptive context chunking, or reinforcement learning guided by downstream response quality.

\subsection{Human expert-based evaluation}

To enable a rigorous comparison of model performance in domain-specific tasks, we conducted an expert-based evaluation of three large language models: GPT-4o, DeepSeek R1, and Emission GPT (our proposed model). The evaluation focused on the models’ capacity to generate technically sound, contextually appropriate, and evidence-based responses to domain-specific regulatory and scientific queries.

Eight PhD-level experts from academia, research institutions, and government agencies independently assessed a set of 15 questions. The questions were categorized into three difficulty levels (Level 1\textendash 3), reflecting increasing cognitive complexity, and covered multiple thematic areas including emission classification, source testing methods, control technologies, inventory compilation, and factor selection. Lower-level factual questions (Level 1) were primarily addressed using standards and technical guidelines, while higher-level reasoning tasks (Level 2 and 3) relied on policy reports, peer-reviewed scientific articles, and academic literature.

\begin{table}[H]
\centering
\caption{Evaluation Dimensions and Scoring Range}
\label{tab:summary_eval}
\begin{tabularx}{\linewidth}{
  >{\centering\arraybackslash}p{3.5cm} 
  >{\centering\arraybackslash}p{3cm} 
  >{\centering\arraybackslash}X
}
\toprule
\textbf{Dimension} & \textbf{Score Range} & \textbf{Core Evaluation Focus} \\
\midrule
Relevance & 0--5 & Alignment with question core; coverage of key terms \\
Accuracy & 0--5 & Factual correctness; logical consistency \\
Specification & 0--5 & Technical terminology; professional depth \\
Citation and Authority & 0--5 & Use of traceable and authoritative sources \\
Overall Impression & 0--5 & Overall clarity, coherence, and informativeness \\
\bottomrule
\end{tabularx}
\end{table}

A five-dimensional scoring rubric was applied, with each dimension rated on a five-point scale: relevance, accuracy, specification, citation and authority, and overall impression. The rubric captured key aspects of response quality, including topic alignment, factual precision, analytical clarity, source reliability, and completeness. Penalties were imposed for major errors such as factual inaccuracies, fabricated references, or irrelevant content. The scoring also accounted for the alignment between question complexity and source types, using standards and guidelines for Level 1 tasks and academic literature for Level 2 and 3.

\begin{figure}[H]
  \centering
  \includegraphics[width=1\textwidth]{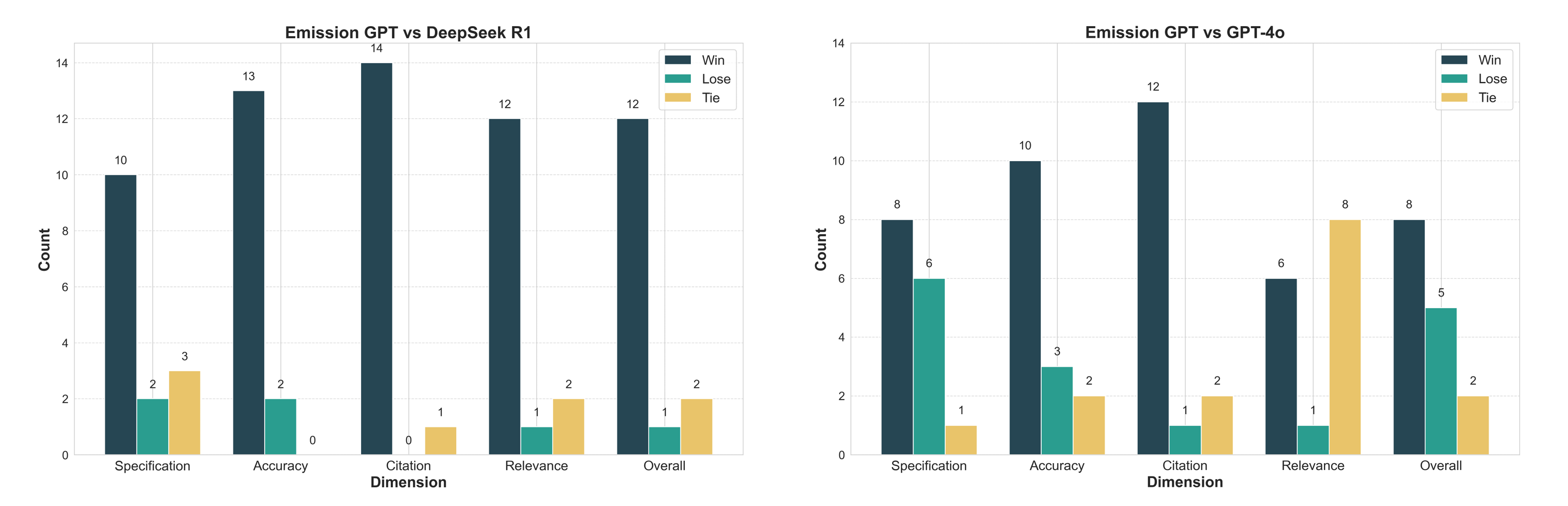}
  \caption{The comparative winning rates of Emission GPT, GPT-4o and DeepSeek R1 across different dimentions}
  \label{fig:expert_framework_bar}
\end{figure}

Results are summarized in Figure~\ref{fig:expert_framework_bar}, which presents the pairwise win-tie-loss comparisons across the five evaluation dimensions between Emission GPT and each of the two baseline models. Emission GPT showed consistent superiority over DeepSeek R1, with a perfect record in citation (14 wins, 1 ties) and clear advantages in relevance (12 wins, 2 tie) and overall impression (12 wins, 2 ties). It also maintained an edge in specification and accuracy. Compared to GPT-4o, Emission GPT secured more wins across all dimensions, particularly in citation and accuracy. GPT-4o was more competitive in specification and richness, reflecting strengths in technical language and elaboration, though sometimes lacking in citation precision. Overall, Emission GPT consistently outperformed both baseline models, particularly in generating accurate, well-supported, and credible responses—key requirements for domain-specific regulatory and scientific tasks

\begin{figure}[H]
  \centering
  \includegraphics[width=1\textwidth]{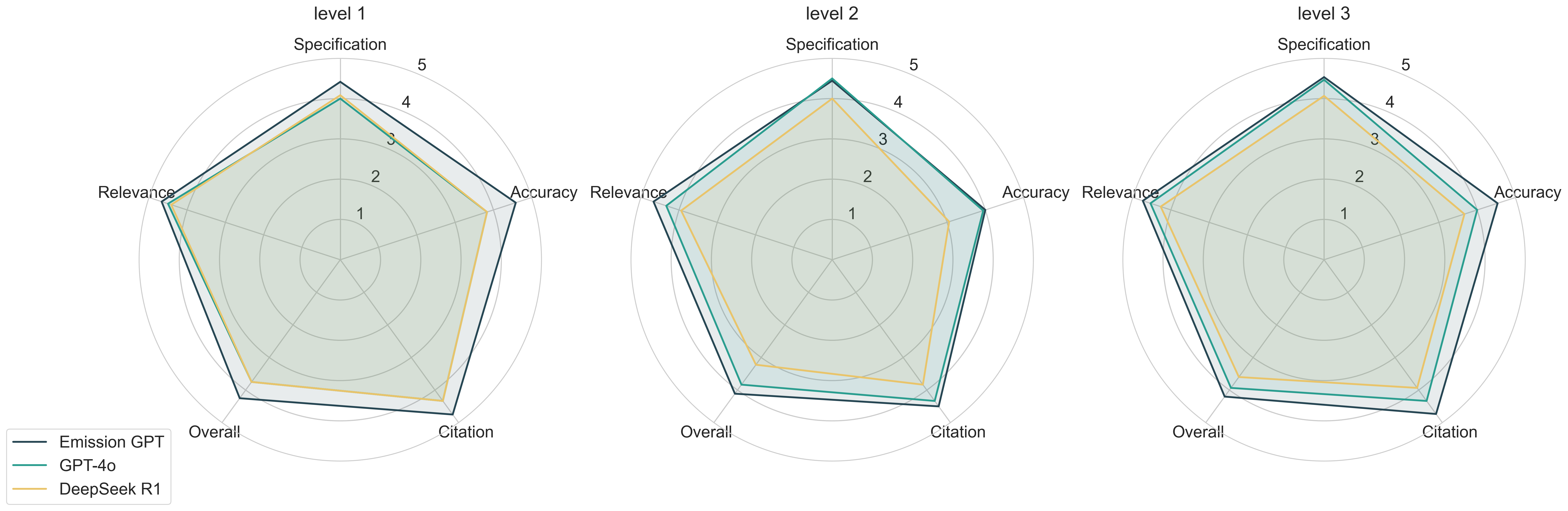}
  \caption{Radar plots of model-specific performance across the five evaluation dimensions for 3 difficulty levels}
  \label{fig:expert_framework_radar}
\end{figure}

As illustrated in Figure~\ref{fig:expert_framework_radar}, Emission GPT achieved the highest or tied-for-highest average scores across all five evaluation dimensions. It outperformed both GPT-4o and DeepSeek R1 in accuracy (4.37), citation (4.66), and relevance (4.69), demonstrating clear advantages in factual correctness, source credibility, and topical alignment. GPT-4o scored slightly higher than DeepSeek R1 in specification (4.32 vs. 4.05), but remained below Emission GPT (4.46). In overall impression, Emission GPT also ranked highest (4.19), suggesting a strong balance of professionalism, credibility, and clarity. GPT-4o showed strength in specification (4.32), slightly above DeepSeek R1 (4.05), but still below Emission GPT. Radar plot comparisons  across difficulty levels showed that all models performed comparably on Level 1 questions. As task complexity increased, Emission GPT’s advantages in accuracy and citation became more evident, while GPT-4o and DeepSeek R1 offered supplementary strengths in specification and richness.

Overall, Emission GPT outperformed both baselines across all evaluation dimensions and difficulty levels. Its strengths in accuracy, citation, and relevance became more prominent as task complexity increased, demonstrating clear advantages in interactive question answering within the complex knowledge system of emission-related domains.

\subsection{Toolchain effectiveness}

To demonstrate the effectiveness of the Emission-GPT toolchain in lowering the technical threshold for emission related data analysis, we present a case study focused on Guangdong Province. By interacting with vehicle emission factor databases and emission inventory data through natural language, the system showcases its ability to support emission analysis without coding or domain expertise. These examples highlight how Emission-GPT lowers technical barriers, improves analytical efficiency, and delivers interpretable, policy-relevant insights.


\subsubsection{A showcase in recommending emission factor}
\label{Emission Factor Recommendation}

Emission-GPT incorporates a dual-stage, query-guided framework for emission factor (EF) recommendation. This functionality enables users to specify structured source attributes—such as emission category, vehicle class, fuel type, standard, and location—via natural language input or API-based parameters. The system interprets user intent, maps it to predefined schema, and executes a tiered retrieval process to ensure both normative compliance and empirical richness.

As illustrated in Figure~\ref{fig:ef_recommendation}, the recommendation workflow begins by retrieving officially published guideline values, when available, based on the user-defined source configuration. These authoritative entries serve as a regulatory reference point and are prioritized for transparency and policy alignment. Subsequently, the system augments this baseline by performing a semantic search within a curated database of peer-reviewed literature and technical reports. Candidate records from the literature are then evaluated and ranked using a composite scoring system that accounts for data representativeness, methodological robustness, sample adequacy, and source credibility.

This two-layer retrieval architecture ensures a balance between regulatory relevance and empirical depth. The guideline-derived entries support conformity with national standards, while the literature-based recommendations enable context-specific refinement when standard values are unavailable, outdated, or incomplete. By providing both sources side by side, Emission-GPT enhances user interpretability and supports traceable, evidence-based selection of emission coefficients for inventory development, air quality modeling, and regulatory reporting.

Overall, this capability addresses key limitations in traditional EF acquisition, including methodological inconsistency, incomplete metadata, and lack of version control. By coupling structured retrieval, semantic reasoning, and quality scoring, Emission-GPT facilitates scientifically rigorous and context-aware EF recommendations, thereby improving transparency and decision confidence in emission science workflows.

\begin{figure}[H]
    \centering
    \includegraphics[width=0.8\textwidth]{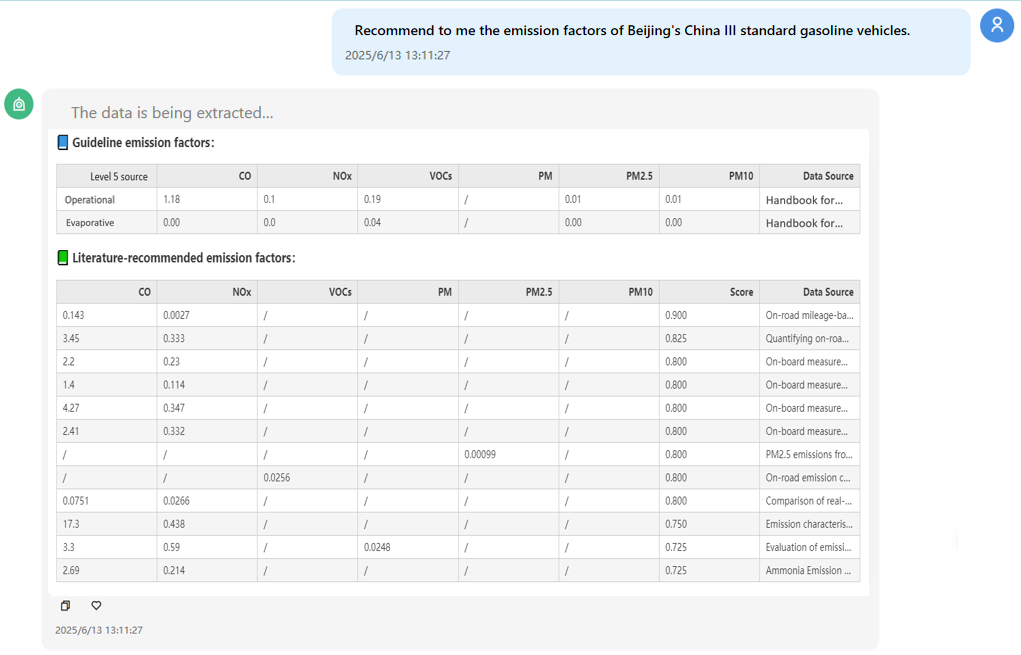}
    \caption{Representative response generated by \textit{Emission-GPT} for emission factor recommendation. The user query specified “gasoline light-duty vehicles under China Stage III standard,” and the system returned a ranked table of candidate emission factors based on multi-dimensional evaluation, including data quality, methodological rigor, and pollutant coverage.}
    \label{fig:ef_recommendation}
\end{figure}

\subsubsection{Emission data retrieval and analysis}
\label{Emission data retrieval and analysis}

The practical effectiveness of Emission-GPT’s toolchain lies in its capacity to transform unstructured expert queries into precise, context-aware analytical operations. Through the integration of modular function libraries, structured prompt engineering, and dynamic function invocation, the system enables seamless interaction with complex emission data sources and knowledge repositories.

A representative set of use cases is illustrated in Figure~\ref{fig:multi_query_results}, which demonstrates Emission-GPT’s capability to address a wide range of emission-related queries with minimal user input. These examples span different pollutant types, spatial and temporal dimensions, and source category breakdowns. In response to natural language queries such as sectoral contributions to NO\textsubscript{x} and CO emissions, annual emission trends of road transport subcategories, or pollutant distributions within mobile sources, the system autonomously identifies appropriate functions, retrieves relevant inventory data, and generates interpretable visual outputs including stacked bar charts and pie charts. The results reveal critical emission patterns: for instance, road mobile sources dominate NO\textsubscript{x} emissions; industrial and mobile sources are primary contributors to CO; and fine-grained breakdowns across transport subcategories provide insights into source-specific mitigation priorities.

This collection of examples underscores the system’s capacity to handle diverse analytic intents, from temporal trend analysis (Figure~\ref{fig:multi_query_results}a) to cross-pollutant source comparison (Figure~\ref{fig:multi_query_results}b), single-pollutant sectoral profiling (Figure~\ref{fig:multi_query_results}c), and intra-sector source composition analysis (Figure~\ref{fig:multi_query_results}d). Such functionality enables domain experts to rapidly iterate through multi-angle diagnostics without manual coding or data preprocessing.

Beyond these illustrative cases, the system supports a broader spectrum of analytical capabilities including inventory validation, regulatory benchmark mapping, and cross-sector comparison. Each function is implemented through a robust orchestration mechanism that ensures schema-compliant input, secure database interaction, and reproducible computation pipelines, thereby ensuring analytical rigor and policy relevance.

Further enhancing robustness, the system is anchored in a prompt engineering framework that constrains the model’s response space, embeds retrieved domain knowledge, and enforces syntactic and semantic consistency in each invocation. This architecture minimizes hallucinated or erroneous outputs, a common limitation in general-purpose language models applied to technical fields.

Collectively, these design features ensure that Emission-GPT serves not only as a conversational interface but also as a trustworthy analytical assistant, capable of executing complex emission-related queries with precision, transparency, and high interpretability.

\begin{figure}[H]
    \centering
    \includegraphics[width=0.8\textwidth]{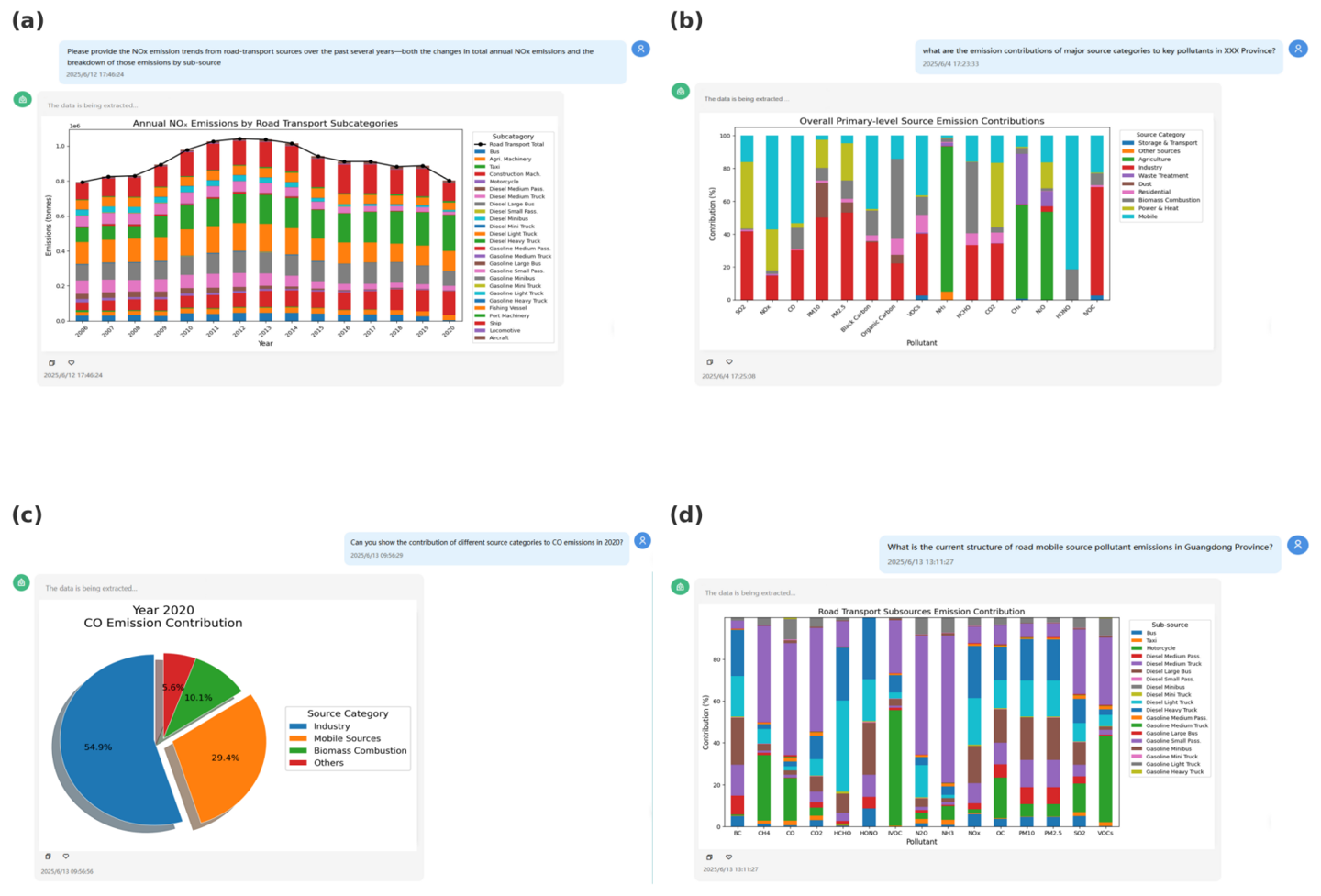}
    \caption{Illustrative responses generated by \textit{Emission-GPT} for different emission analysis queries. (a) Annual NO\textsubscript{x} emissions from road transport subcategories. (b) Cross-pollutant emission contributions by major source categories. (c) CO emission contribution by sector in 2020. (d) Road transport sub-source contributions by pollutant.}
    \label{fig:multi_query_results}
\end{figure}

\subsection{Limitations and future work}
While our study demonstrates the potential of multi-agents LLMs to support atmospheric emission research and policy through intelligent document analysis and question answering, there remain several areas for further enhancement. 

First, the current system mainly supports knowledge retrieval and synthesis from textual documents. However, real-world emission research often relies on diverse information types, including structured inventory datasets, numerical time series, and geospatial imagery. Therefore, a future priority is to extend Emission-GPT into a modular agent-based architecture, incorporating specialized agents for emission inventory compilation, data quality assessment, and near-real-time emission estimation, among others.

Second, while the RAG architecture enables effective retrieval, it lacks the capacity to reason over structured knowledge and complex relationships\textendash for example, linking emission source classifications, pollutant types, and regulatory standards. To address this, we plan to integrate domain-specific knowledge base into the knowledge graph to enable systematic reasoning across interconnected entities, such as matching emission factors with source categories or connecting standards to specific pollutants.

Third, although the manual document collection was performed to provide broad coverage, limited and time-consuming manual access to proprietary standards and recent peer-reviewed publications may have led to information gaps. To improve coverage and timeliness, we will implement an automated updating pipeline to continuously ingest new regulatory documents, standards, and research papers.

Finally, Emission-GPT currently cannot process visual content such as scanned documents, figures, or embedded tables. This limitation excludes some technical materials that contain important information in figures and images. To overcome this, we are integrating OCR-based parsing modules to enable structured information extraction from diverse document formats. As emission science evolves toward more integrated, cross-sectoral analyses, Emission-GPT will continue to evolve as a robust platform for emission knowledge retrievals, emission inventory compilation, policy making, and real-world decision support in the field of atmospheric emission.

\newpage

\appendix
\section{Knowledge Base Composition and Topical Coverage}
\label{Knowledge Base Composition and Topical Coverage}

The atmospheric emission knowledge database integrates a comprehensive collection of resources, classified systematically according to type, publication year, issuing institutions, and core content.

For the journal article subset, the database employs detailed classification dimensions including research region, emission sources, pollutants, and thematic categories. Region-specific entries cover 40 geographic expressions. Studies span national, sub-national, and basin-level scales, with frequently represented regions including China, the United States, European countries and the Yellow River Basin. The predominant emission sectors of research include Industrial Sources (50.4\%), Agricultural Sources (46.2\%), and Biomass Burning (38.4\%), reflecting significant attention to large-scale anthropogenic emissions. Major pollutants covered are $CO_2$ (41.1\%), $NO_X$ (25.2\%), and $PM_{2.5}$ (19.0\%), emphasizing both greenhouse gases and key air pollutants. Key thematic areas covered extensively are emission inventories and trend analyses (55.7\%), emissi on modeling (46.8\%), source apportionment studies (45.7\%), comprehensive reviews (41.3\%), and emission factor management (17.3\%).

The report data covers an extensive period from 1972 to 2025 and includes 809 documents from diverse issuing institutions. The primary sources are the U.S. EPA, which accounts for 76.0\% of the documents, followed by China's Ministry of Ecology and Environment and its affiliates at 16.7\%, and international agencies like the IPCC at 3.7\%. Document types are varied, with compiled technical volumes (26.0\%) and formal reports (21.9\%) being most common, followed by regulatory and guidance documents (12.2\%), fact sheets (9.9\%), environmental bulletins (7.2\%), and references (6.6\%). These reports provide historical context and practical guidance for policy formulation and environmental management

The collection of scholarly books published from 2006 to 2024 focuses on key areas such as emission inventory methodologies, atmospheric chemistry, air quality modeling, and health impacts associated with pollution. Published by reputable institutions like Science Press and China Environmental Science Press, these monographs serve as authoritative references, providing comprehensive theoretical frameworks and methodological guidelines.

\section{Toolchain Effectiveness}
\label{Toolchain Effectiveness}

\subsection{Example of emission data retrieval and analysis}
\label{Example of emission data retrieval and analysis}

The practical effectiveness of Emission-GPT’s toolchain lies in its capacity to transform unstructured expert queries into precise, context-aware analytical operations. Through the integration of modular function libraries, structured prompt engineering, and dynamic function invocation, the system enables seamless interaction with complex emission data sources and knowledge repositories.

A representative set of use cases is illustrated in Figure~\ref{fig:multi_query_results}, which demonstrates Emission-GPT’s capability to address a wide range of emission-related queries with minimal user input. These examples span different pollutant types, spatial and temporal dimensions, and source category breakdowns. In response to natural language queries such as sectoral contributions to NO\textsubscript{x} and CO emissions, annual emission trends of road transport subcategories, or pollutant distributions within mobile sources, the system autonomously identifies appropriate functions, retrieves relevant inventory data, and generates interpretable visual outputs including stacked bar charts and pie charts. The results reveal critical emission patterns: for instance, road mobile sources dominate NO\textsubscript{x} emissions; industrial and mobile sources are primary contributors to CO; and fine-grained breakdowns across transport subcategories provide insights into source-specific mitigation priorities.

This collection of examples underscores the system’s capacity to handle diverse analytic intents, from temporal trend analysis (Figure~\ref{fig:multi_query_results}a) to cross-pollutant source comparison (Figure~\ref{fig:multi_query_results}b), single-pollutant sectoral profiling (Figure~\ref{fig:multi_query_results}c), and intra-sector source composition analysis (Figure~\ref{fig:multi_query_results}d). Such functionality enables domain experts to rapidly iterate through multi-angle diagnostics without manual coding or data preprocessing.

Beyond these illustrative cases, the system supports a broader spectrum of analytical capabilities including inventory validation, regulatory benchmark mapping, and cross-sector comparison. Each function is implemented through a robust orchestration mechanism that ensures schema-compliant input, secure database interaction, and reproducible computation pipelines, thereby ensuring analytical rigor and policy relevance.

Further enhancing robustness, the system is anchored in a prompt engineering framework that constrains the model’s response space, embeds retrieved domain knowledge, and enforces syntactic and semantic consistency in each invocation. This architecture minimizes hallucinated or erroneous outputs, a common limitation in general-purpose language models applied to technical fields.

Collectively, these design features ensure that Emission-GPT serves not only as a conversational interface but also as a trustworthy analytical assistant, capable of executing complex emission-related queries with precision, transparency, and high interpretability.

\subsection{Example of emission factor recommendation}
\label{Example of emission factor recommendation}

In addition to data-centric emission diagnostics, Emission-GPT incorporates a dual-stage, query-guided framework for emission factor (EF) recommendation. This functionality enables users to specify structured source attributes—such as emission category, vehicle class, fuel type, standard, and location—via natural language input or API-based parameters. The system interprets user intent, maps it to predefined schema, and executes a tiered retrieval process to ensure both normative compliance and empirical richness.

As illustrated in Figure~\ref{fig:ef_recommendation}, the recommendation workflow begins by retrieving officially published guideline values, when available, based on the user-defined source configuration. These authoritative entries serve as a regulatory reference point and are prioritized for transparency and policy alignment. Subsequently, the system augments this baseline by performing a semantic search within a curated database of peer-reviewed literature and technical reports. Candidate records from the literature are then evaluated and ranked using a composite scoring system that accounts for data representativeness, methodological robustness, sample adequacy, and source credibility.

This two-layer retrieval architecture ensures a balance between regulatory relevance and empirical depth. The guideline-derived entries support conformity with national standards, while the literature-based recommendations enable context-specific refinement when standard values are unavailable, outdated, or incomplete. By providing both sources side by side, Emission-GPT enhances user interpretability and supports traceable, evidence-based selection of emission coefficients for inventory development, air quality modeling, and regulatory reporting.

Overall, this capability addresses key limitations in traditional EF acquisition, including methodological inconsistency, incomplete metadata, and lack of version control. By coupling structured retrieval, semantic reasoning, and quality scoring, Emission-GPT facilitates scientifically rigorous and context-aware EF recommendations, thereby improving transparency and decision confidence in emission science workflows.

\begin{figure}[H]
    \centering
    \includegraphics[width=0.8\textwidth]{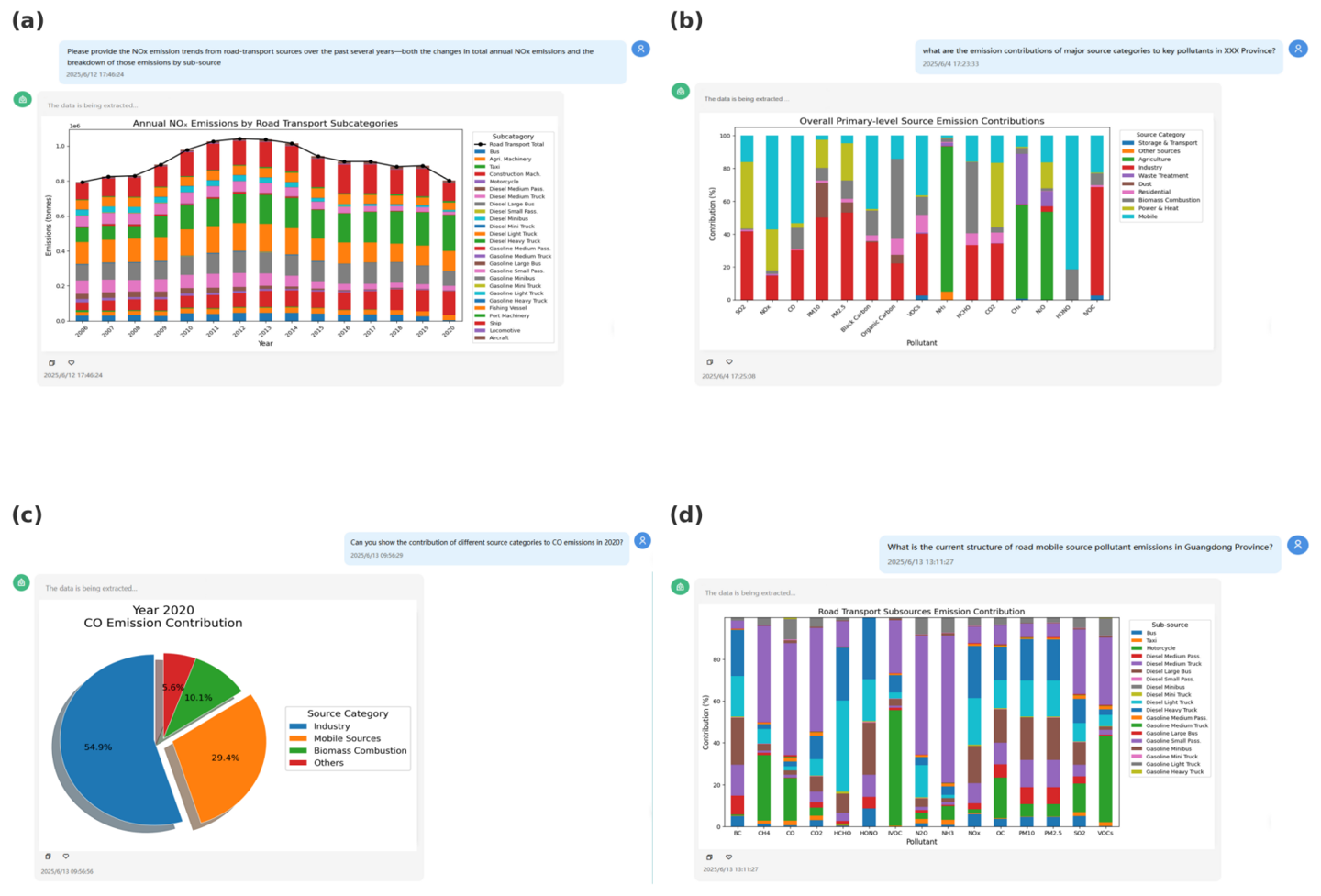}
    \caption{Illustrative responses generated by \textit{Emission-GPT} for different emission analysis queries. (a) Annual NO\textsubscript{x} emissions from road transport subcategories. (b) Cross-pollutant emission contributions by major source categories. (c) CO emission contribution by sector in 2020. (d) Road transport sub-source contributions by pollutant.}
    \label{fig:multi_query_results}
\end{figure}

\begin{figure}[H]
    \centering
    \includegraphics[width=\textwidth]{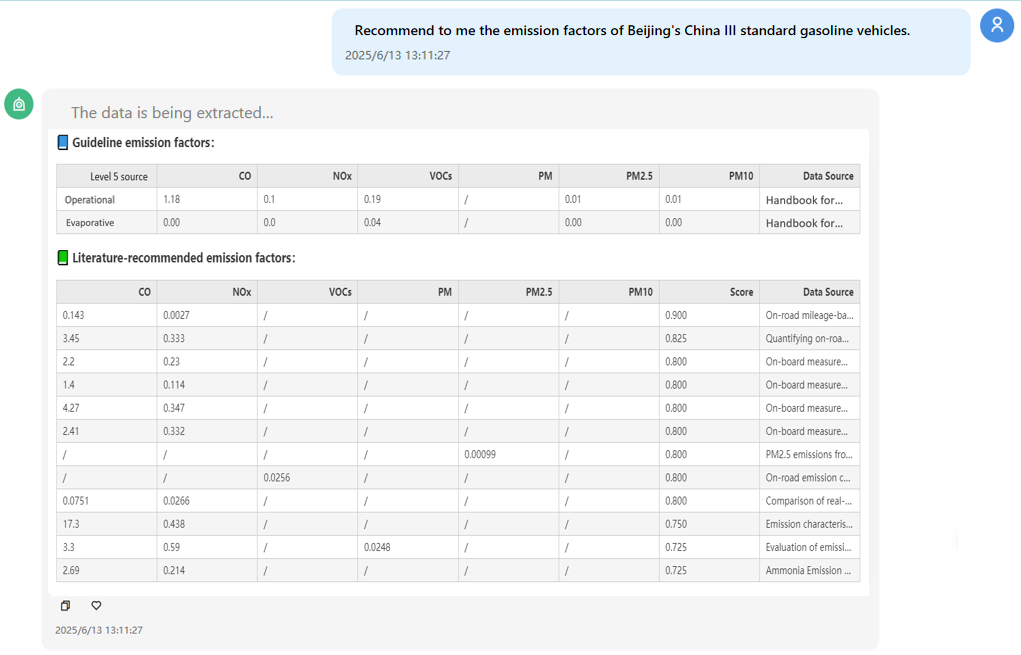}
    \caption{Representative response generated by \textit{Emission-GPT} for emission factor recommendation. The user query specified “gasoline light-duty vehicles under China Stage III standard,” and the system returned a ranked table of candidate emission factors based on multi-dimensional evaluation, including data quality, methodological rigor, and pollutant coverage.}
    \label{fig:ef_recommendation}
\end{figure}

\begin{table}[htbp]
\centering
\caption{Grading criteria for emission data quality}\label{table: EF_recommendation}
\renewcommand{\arraystretch}{1.5}
\begin{tabular}{>{\centering\arraybackslash}p{0.8cm}  
                >{\raggedright\arraybackslash}p{3.5cm}   
                >{\raggedright\arraybackslash}p{3.5cm}   
                >{\raggedright\arraybackslash}p{3cm} 
                >{\raggedright\arraybackslash}p{3cm}}
\toprule
\textbf{Grade} & \textbf{Data representativeness} & \textbf{Methodological reliability} & \textbf{Sample representativeness} & \textbf{Data authority} \\
\midrule
A & Data originates from the same study region, testing or publication conducted within the past 5 years & Uses standardized methods validated by sufficient data, control technologies are comprehensively considered & Sample size is large, representativeness is good, emission characteristics are well reflected & Published on platforms with rigorous peer review, recognized institutional authority \\
B & From a similar region with one-level scale difference, testing or publication within the past 10 years & Uses reliable but not fully standardized methods, control technologies are partially considered, some uncertainty exists & Sample size is moderate, representativeness is acceptable, some limitations exist & From sources with formal publication qualifications, relatively limited academic influence \\
C & From a similar region with two-level scale difference, testing or publication within the past 15 years & Method lacks validation, key components are missing, control technologies are insufficiently described & Sample size is small, representativeness is weak, only partial situations reflected & Source lacks clear review process, acceptance standards are uncertain, credibility is questionable \\
D & From unrelated or unknown region, testing time is unknown or exceeds 15 years & Method is unacceptable or undocumented, control conditions are unclear, technical explanations are missing & Sample information not provided, data is incomplete, representativeness cannot be judged & Source is unverifiable, data traceability is absent \\
\bottomrule
\end{tabular}
\end{table}

\section{Q\&A Comparison for Emission GPT, GPT-4o, and DeepSeek R1}

\subsection*{Q1: How to derive black carbon emission factors from different anthropogenic sources? Provide answers with citation.}

\subsubsection*{Answer Produced by Emission GPT}

Black carbon (BC) emission factors from different anthropogenic sources can be derived using particulate matter (PM) emission factors and the black carbon fraction (BCF). Below is a detailed explanation of how these factors are estimated for various sources:

For diesel engines, the BC emission factor can be calculated using the formula:
\[
EF_{BC} = EF_{PM_{2.5}} \times BCF
\]
Where $EF_{BC}$ is the black carbon emission factor (g/hp·h), $EF_{PM_{2.5}}$ is the emission factor for PM$_{2.5}$ (g/hp·h), and $BCF$ is the black carbon fraction (gBC/gPM$_{2.5}$). The black carbon fraction for diesel engines is typically 0.77~\cite{ref1}.

For gasoline engines, the black carbon fraction is significantly lower compared to diesel engines. The BCF for gasoline engines is approximately $EF_{BC0}$~\cite{ref1}. 

Combustion sources such as coal burning, biomass burning, and fossil fuel combustion contribute significantly to BC emissions. In coal-fired power plants, the BC emission factor typically ranges from $EF_{BC1}$ to $EF_{BC2}$ grams per kilogram of coal burned~\cite{ref2}.

Biomass burning emissions vary by category. For instance, agricultural waste burning may emit around $EF_{BC4}$ to $EF_{BC5}$ grams of BC per kilogram of biomass burned~\cite{ref3}.

Industrial high-temperature combustion (e.g., in steel production) may yield BC emission factors ranging from $EF_{BC6}$ to $EF_{BC7}$~\cite{ref5}.

For heavy-duty diesel vehicles, BC emission factors typically range from $EF_{BC8}$ to $EF_{BC9}$ grams per kilometer~\cite{ref6}.

\subsubsection*{Answer Produced by GPT-4o}

Deriving black carbon (BC) emission factors from different anthropogenic sources involves a combination of experimental measurement, mass balance estimation, and empirical modeling. The method depends on the type of source (e.g., combustion device, vehicle, industrial process), fuel type, and available data. Here's a step-by-step overview:

\textbf{1. Direct Measurement Approach (Preferred)}  
\textbf{Equation:}  
\[
\text{EF}_{BC} = \frac{M_{BC}}{M_{fuel}} \quad (\text{units: g BC / kg fuel})
\]  
Where $M_{BC}$ is the mass of black carbon emitted (g), and $M_{fuel}$ is the mass of fuel burned (kg).

\textbf{Required Tools:} Gravimetric filter samplers, thermal-optical analyzers, fuel flow meters.  
\textbf{Example:} Use NIOSH 5040 protocol to analyze diesel engine exhaust.

\textbf{2. Carbon Mass Balance Method}  
\[
\text{EF}_{BC} = C_f \times \frac{[BC]}{[CO_2] + [CO] + [BC] + [OC]}
\]  
Where $C_f$ is the carbon content per kg fuel, and $[X]$ are emission measurements.

\textbf{3. Inventory-Based (Top-Down)}  
\[
\text{EF}_{BC} = \frac{E_{BC}}{A}
\]  
Where $E_{BC}$ is total BC emissions and $A$ is activity (e.g., fuel burned).

\textbf{4. Scaling from PM$_{2.5}$:}  
\[
\text{EF}_{BC} = \text{EF}_{PM_{2.5}} \times f_{BC}
\]  
Example: If EF$_{PM_{2.5}}$ = 2.0 g/kg, and $f_{BC}$ = 0.6, then EF$_{BC}$ = 1.2 g/kg.

\textbf{Key References:}  
Bond et al. (2004, 2013), USEPA AP-42, IPCC 2019, EMEP/EEA (2019), Andreae \& Merlet (2001), Zhang et al. (2009).

\textbf{Tips:} Match source category precisely, consider combustion conditions, adjust for emission controls, and use local data where possible.

\subsubsection*{Answer Produced by DeepSeek R1}

Black Carbon (BC) is a short-lived climate pollutant emitted from incomplete combustion. To quantify BC emissions, emission factors (EFs)—the amount of BC released per unit activity—must be derived. Below is a structured method:

\textbf{Step 1: Source Classification}  
Identify source: transport (diesel), industry (coal plants), residential (stoves), agriculture (crop burning).

\textbf{Step 2: Data Collection}  
(A) \textit{Direct Measurements:} Aethalometers, thermal-optical methods (e.g., IMPROVE protocol).  
(B) \textit{Literature:} EPA AP-42, EMEP/EEA.  
Example values:
\begin{itemize}
  \item Diesel trucks: 0.1–0.3 g/kg (Bond et al. 2013)
  \item Coal boilers: 0.5–2.0 g/kg (IPCC 2019)
  \item Wood: 0.3–1.5 g/kg (Andreae 2019)
\end{itemize}

\textbf{Step 3: Activity Data}  
Fuel burned, operating time, vehicle kilometers traveled.

\textbf{Step 4: Uncertainty Analysis}  
Use Monte Carlo simulations; uncertainty up to ±30\%.

\textbf{Methodologies:}  
(A) \textit{Lab Testing:} Thermal-optical analysis, gravimetric filters.  
(B) \textit{Field:} Portable aethalometers (AE33), brick kiln tests.

\textbf{Example EF Ranges:}
\begin{itemize}
  \item Diesel vehicles: 0.1–0.5 g/kg (up to 1.5 for older)  
  \item Residential coal: 0.5–3.0 g/kg  
  \item Biomass: 0.3–1.5 g/kg  
  \item Ships (HFO): 0.1–0.4 g/kg  
  \item Brick kilns: 0.8–2.5 g/kg
\end{itemize}

\textbf{Key References:}  
Bond et al. (2013), IPCC (2019), EMEP/EEA (2023), Andreae (2019), ICCT (2020), Weyant et al. (2019)

\textbf{Challenges:}  
Lack of local data, real-world vs. lab gap, regional variation, satellite-based EFs (e.g., GFED4, EDGAR).

\subsection*{Q2: In current wildfire emission estimation methods, how is PM-2.5 calculated from PM-10, and what activity and fuel-related factors are involved in the estimation?}

\subsubsection*{Anwer Poruced by Emission GPT}

In current wildfire emission estimation methods, PM-2.5 is often calculated from PM-10 by applying a size distribution factor based on measured size distributions. Specifically, PM-2.5 is estimated to be 0.9 times the PM-10 value, assuming a consistent proportion of fine particles within the total particulate matter emitted \cite{ref1}. This approach simplifies the estimation process but relies heavily on the accuracy and representativeness of the assumed size distribution.

The estimation of PM-2.5 emissions from wildfires involves several activity and fuel-related factors. One key variable is the activity factor, which is expressed as the sum of land acres burned at the fuel category level for the wildfires. This data is typically sourced from agencies such as the National Forest Service and the U.S. Department of the Interior, providing spatially resolved information on the extent of burning \cite{ref1}.

Another critical component is the fuel loading factor, which represents the average amount of fuel available for combustion per unit area. This factor is derived from databases like AP-42 and other specific projects that provide regionally specific values for different types of vegetation and land cover \cite{ref1}. Fuel loading varies significantly depending on the ecosystem, with forests, grasslands, and shrublands having distinct characteristics that influence the quantity of emissions produced during a fire event.

Additional parameters include the combustion efficiency and emission factors, which are influenced by the type of fuel (e.g., hardwood, softwood, grasses), its moisture content, and the intensity and duration of the fire. These factors determine how much of the fuel is actually consumed and how the emissions are distributed across different particle sizes, including PM-2.5 \cite{ref3}.

Finally, uncertainties in these methods arise due to variability in fire behavior, differences in fuel characteristics, and limitations in spatial and temporal resolution of input data. As noted in some studies, future improvements may involve using chemistry transport modeling to better quantify the contribution of wildfire emissions to PM-2.5 levels \cite{ref2}.

\subsubsection*{Answer Produced by GPT-4o}

1. PM2.5 Estimation  
In wildfire emission inventories, PM2.5 is commonly derived from PM10 by applying a size-distribution factor. A frequently used assumption is:  
PM2.5 \( \approx 0.9 \times \) PM10, based on empirical measurements showing that a large portion of wildfire-generated particles fall within the fine fraction.

2. Activity Factor: Burned Area  
The activity factor represents the total land area burned and is typically categorized by fuel type. Data are sourced from federal agencies such as the U.S. Forest Service, Bureau of Indian Affairs, Bureau of Land Management, and National Park Service.

3. Fuel Loading Factor  
The fuel loading factor quantifies the average biomass (e.g., wood, litter, shrubs) per unit area. Fuel loading values are often taken at the state level using empirical estimates documented in EPA AP 42 or derived from region-specific studies.

4. Emission Factor Equation  
Wildfire emissions are calculated using a well-established formula:
\[
E = A \times FL \times CC \times EF
\]
Where A = area burned, FL = fuel loading, CC = combustion completeness, and EF = pollutant-specific emission factor (e.g., PM2.5).

Key References:  
\begin{itemize}
  \item Urbanski et al. (2014)
  \item EPA NEI and AP 42 manuals (1999--2002)
  \item AP 42 Chapter 13.1 (EPA)
  \item Field studies (e.g., Jenkins et al.)
\end{itemize}

This methodology---combining area burned (A), fuel loading (FL), combustion completeness (CC), and emission factor (EF), with the application of a 0.9 multiplier to estimate PM2.5 from PM10---forms the backbone of current wildfire emission inventories.

\subsubsection*{Answer Produced by DeepSeek R1}

In current wildfire emission estimation methods, PM2.5 is often estimated from PM10 using mass ratios derived from field measurements or modeling studies. The relationship between PM2.5 and PM10 depends on combustion conditions, fuel type, and fire behavior.

1. PM2.5/PM10 Ratio in Wildfire Emissions
\begin{itemize}
  \item PM2.5 typically constitutes 60--90\% of PM10 in wildfire smoke, depending on fuel type and combustion phase (flaming vs. smoldering).
  \item Flaming combustion tends to produce finer particles (higher PM2.5/PM10 ratio), while smoldering combustion generates more coarse particles (lower PM2.5/PM10 ratio).
  \item Default ratios are often used in emission models when direct PM2.5 measurements are unavailable.
\end{itemize}

2. Activity and Fuel-Related Factors Affecting PM Emissions
Wildfire PM emissions depend on:

\textbf{Fuel Characteristics}
\begin{itemize}
  \item Fuel type (e.g., grasses, shrubs, forests): Fine fuels (grass) burn more completely, producing higher PM2.5 fractions.
  \item Fuel moisture: Drier fuels burn more efficiently, influencing particle size distribution.
  \item Fuel loading: Higher fuel density leads to greater total emissions.
\end{itemize}

\textbf{Fire Behavior and Combustion Phase}
\begin{itemize}
  \item Flaming vs. smoldering: Smoldering fires emit more PM2.5 relative to PM10.
  \item Fire intensity: High-intensity fires may produce more coarse PM due to lofting of ash.
\end{itemize}

\textbf{Meteorological and Environmental Conditions}
\begin{itemize}
  \item Wind speed: Influences particle dispersion and secondary aerosol formation.
  \item Atmospheric stability: Affects how PM is transported and aged.
\end{itemize}

Key Literature Supporting These Methods:
\begin{itemize}
  \item Andreae (2019). \textit{Atmospheric Chemistry and Physics}, 19(13), 8523--8546. DOI: \url{https://doi.org/10.5194/acp-19-8523-2019}
  \item Urbanski et al. (2018). \textit{Earth System Science Data}, 10(4), 2241--2274. DOI: \url{https://doi.org/10.5194/essd-10-2241-2018}
  \item Reid et al. (2005). \textit{Atmospheric Chemistry and Physics}, 5(3), 799--825. DOI: \url{https://doi.org/10.5194/acp-5-799-2005}
  \item Wiedinmyer et al. (2011). \textit{Geoscientific Model Development}, 4(3), 625--641. DOI: \url{https://doi.org/10.5194/gmd-4-625-2011}
  \item U.S. EPA (2022). SPECIATE Database. \url{https://www.epa.gov/air-emissions-modeling/speciate}
\end{itemize}

\textbf{Conclusion:} PM2.5 is typically derived from PM10 using empirical ratios, which vary by fuel type and combustion conditions. Key factors influencing emissions include fuel properties, fire behavior, and environmental conditions. The cited literature provides further details on emission estimation methodologies.

\subsection*{Q3: Provide representative case studies related to ammonia (NH$_3$) emissions from major anthropogenic sources with citations}

\subsubsection*{Anwer Poruced by Emission GPT}

Ammonia ($\mathrm{NH}_3$) emissions from anthropogenic sources are a significant contributor to air pollution and secondary aerosol formation. Below are representative case studies and datasets that highlight the magnitude, sources, and spatial distribution of ammonia emissions.

\textbf{Case Study 1: Global Ammonia Emissions (2006)}  
A global estimate of ammonia emissions was provided in the literature, categorizing both natural and anthropogenic sources. The total global ammonia emissions were estimated at \textbf{45 Tg/year}, with the breakdown as follows:
\begin{itemize}
  \item \textbf{Anthropogenic sources}: 30.4 Tg/year
  \begin{itemize}
    \item Livestock (e.g., horses, cattle, sheep): ~18.7 Tg/year
    \item Fertilizers: 6.4 Tg/year
    \item Biomass burning: 2.0 Tg/year
  \end{itemize}
  \item \textbf{Natural sources}: 14.6 Tg/year
  \begin{itemize}
    \item Wildlife: 2.5 Tg/year
    \item Plants: 5.1 Tg/year
    \item Oceans: 7.0 Tg/year
  \end{itemize}
\end{itemize}
This dataset underscores the dominance of anthropogenic sources, particularly livestock farming and fertilizer use [1].

\textbf{Case Study 2: Jiangsu Province, China (2006--2014)}  
In Jiangsu Province, ammonia emissions were studied over the period 2006--2014. The total emissions increased from \textbf{142,000 tons in 2006} to \textbf{165,000 tons in 2014}, reflecting growing agricultural and industrial activities. Key findings include:
\begin{itemize}
  \item \textbf{Fertilizer application}: ~50\% of total emissions
  \item \textbf{Livestock farming}: ~40\% of total emissions
\end{itemize}
Spatially, higher emissions were concentrated in southern Jiangsu, where economic activities are more intensive. The study also revealed that urban areas had higher emission intensities due to combined contributions from agriculture, industry, and transportation [2].

\textbf{Case Study 3: Yangtze River Delta (YRD), China}  
The YRD region is one of the most economically developed areas in China, with significant ammonia emissions. A high-resolution ammonia emission inventory for the YRD was developed, showing that:
\begin{itemize}
  \item \textbf{Agricultural sources} (livestock farming and fertilizer use) accounted for approximately \textbf{93.7\%} of total emissions.
  \item Non-agricultural sources, such as industrial production, human emissions, biomass burning, and traffic, contributed the remaining \textbf{6.3\%}.
  \item Spatially, emissions were highest in densely populated and industrialized cities like Shanghai, Nanjing, and Hangzhou.
\end{itemize}
This case highlights the importance of agricultural activities in ammonia emissions in the YRD region [3].

\textbf{Case Study 4: United States (EPA Data, 2011)}  
In the United States, ammonia emissions were estimated using county-level data for the year 2011. Key findings include:
\begin{itemize}
  \item \textbf{Livestock farming}: Dominant source, accounting for ~80\% of total ammonia emissions.
  \item \textbf{Synthetic fertilizers}: Significant contributor, especially in agricultural states like Iowa and Illinois.
  \item \textbf{Other sources}: Minor contributions from wastewater treatment, human activities, and industrial processes.
\end{itemize}
For instance, in Iowa, ammonia emissions from livestock farming were estimated at \textbf{120,000 tons}, primarily driven by large-scale pig and cattle farming operations [4].

\textbf{Case Study 5: European Union (EEA Data, 2011)}  
The European Environment Agency (EEA) reported ammonia emissions for its member countries in 2011. Total emissions were estimated at \textbf{428 kt}, with agricultural sources contributing \textbf{93.7\%} of the total. Key contributors included:
\begin{itemize}
  \item \textbf{Dairy farming}: Major contributor in countries like the Netherlands and Denmark.
  \item \textbf{Crop fertilization}: Significant in cereal-producing regions like France and Germany.
  \item \textbf{Biomass burning}: Minor but notable in Eastern European countries.
\end{itemize}
For example, the Netherlands reported ammonia emissions of \textbf{140,000 tons in 2015}, with dairy farming contributing ~60\% of the total [5].

\textbf{Conclusion}  
These case studies and datasets demonstrate the variability in ammonia emissions across different regions and sectors. Agricultural activities, particularly livestock farming and fertilizer application, dominate ammonia emissions globally. However, regional differences in economic structure, population density, and environmental policies influence the magnitude and composition of emissions. Understanding these patterns is crucial for developing effective mitigation strategies.

\subsubsection*{Anwer Poruced by GPT-4o}

Here are representative case studies on ammonia (NH$_3$) emissions from major anthropogenic sources, each reflecting distinct emission sectors such as agriculture, waste management, and industrial activity:

\textbf{1. Agricultural Sources – Livestock and Fertilizer Application}  
Case Study: European Ammonia Emissions from Livestock Housing and Manure Management  
• Study: [Hutchings et al., 2001]  
• Location: Denmark  
• Summary: This study quantified NH$_3$ emissions from various livestock systems including pig and cattle housing, manure storage, and field application. It highlighted the high emission potential during field spreading of slurry and urea-based fertilizers.  
• Findings:  
  o Housing: up to 40\% of total NH$_3$ emissions.  
  o Manure spreading: up to 50\% of total emissions.  
  o Mitigation potential using low-emission spreading techniques like injection or band spreading.  
• Citation: Hutchings, N. J., Sommer, S. G., Andersen, J. M., \& Asman, W. A. H. (2001). A detailed ammonia emission inventory for Denmark. \textit{Atmospheric Environment}, 35(11), 1959–1968. \url{https://doi.org/10.1016/S1352-2310(00)00457-1}

\textbf{2. Fertilizer-Driven Emissions – Cropland in Asia}  
Case Study: Ammonia Emissions from Rice-Wheat Systems in the Indo-Gangetic Plains  
• Study: [Bouwman et al., 2002]  
• Location: India  
• Summary: NH$_3$ emissions from nitrogen fertilizer use in rice-wheat rotation systems were assessed using both empirical field data and modeling. Urea-based fertilizers were the dominant emission source.  
• Findings:  
  o Emission factors varied with temperature, soil pH, and fertilizer placement.  
  o Seasonal peaks in emissions aligned with fertilization and irrigation events.  
• Citation: Bouwman, A. F., Boumans, L. J. M., \& Batjes, N. H. (2002). Emissions of N$_2$O and NO from fertilized fields: Summary of available measurement data. \textit{Global Biogeochemical Cycles}, 16(4). \url{https://doi.org/10.1029/2001GB001811}

\textbf{3. Waste Management – Municipal Wastewater and Landfills}  
Case Study: NH$_3$ Emissions from Urban Wastewater Treatment Plants  
• Study: [Batstone et al., 2015]  
• Location: Australia  
• Summary: This research evaluated ammonia volatilization during sludge digestion and aeration processes in urban WWTPs. Significant NH$_3$ release was observed in sludge drying beds and aeration tanks.  
• Findings:  
  o NH$_3$ losses accounted for ~15\% of total nitrogen processed.  
  o Seasonal and temperature-dependent emission variation.  
• Citation: Batstone, D. J., Hülsen, T., Mehta, C. M., \& Keller, J. (2015). Platforms for energy and nutrient recovery from domestic wastewater: A review. \textit{Chemosphere}, 140, 2–11. \url{https://doi.org/10.1016/j.chemosphere.2014.10.021}

\textbf{4. Industrial Processes – Fertilizer and Chemical Manufacturing}  
Case Study: Ammonia Emissions from Nitric Acid and Urea Plants in China  
• Study: [Kuhn et al., 2020]  
• Location: China  
• Summary: NH$_3$ emissions were monitored using mobile differential optical absorption spectroscopy (DOAS) at several fertilizer plants. This study provided near-field data on fugitive and stack emissions from large-scale facilities.  
• Findings:  
  o Stack emissions from urea plants reached 15–30 kg NH$_3$/hour.  
  o Fugitive emissions from storage and handling were also significant.  
• Citation: Kuhn, T., et al. (2020). Mobile DOAS measurements of ammonia around Chinese fertilizer plants. \textit{Atmospheric Chemistry and Physics}, 20(3), 1601–1614. \url{https://doi.org/10.5194/acp-20-1601-2020}

\textbf{5. Biomass Burning – Agricultural Residue and Forest Fires}  
Case Study: Ammonia Emissions from Agricultural Residue Burning in Southeast Asia  
• Study: [Streets et al., 2003]  
• Location: Southeast Asia  
• Summary: Emission inventories estimated NH$_3$ released during post-harvest burning of rice straw and other residues. The study used satellite-derived activity data and emission factors to estimate regional budgets.  
• Findings:  
  o NH$_3$ emission factor for rice straw burning: ~1.4 g/kg dry matter.  
  o Contributed significantly to regional NH$_3$ concentrations during dry season.  
• Citation: Streets, D. G., et al. (2003). An inventory of gaseous and primary aerosol emissions in Asia in the year 2000. \textit{Journal of Geophysical Research: Atmospheres}, 108(D21). \url{https://doi.org/10.1029/2002JD003093}

\subsubsection*{Anwer Poruced by DeepSeek R1}

Here are representative case studies related to ammonia (NH$_3$) emissions from major anthropogenic sources, along with citations:

\textbf{1. Agriculture (Livestock \& Fertilizer Use)}  
Case Study: NH$_3$ emissions from intensive dairy farming in the Netherlands  
• Findings: Livestock manure management (housing, storage, and application) contributed ~80\% of national NH$_3$ emissions. Slurry injection reduced emissions by 50\% compared to surface application.  
• Citation: Velthof et al. (2012). Ammonia emissions from dairy farming systems in the Netherlands. \textit{Nutrient Cycling in Agroecosystems}, 93(1), 23–40. \url{https://doi.org/10.1007/s10705-011-9433-3}

\textbf{2. Fertilizer Application (Croplands)}  
Case Study: NH$_3$ volatilization from urea fertilization in China  
• Findings: Urea application in Chinese wheat fields led to NH$_3$ losses of 15–40\% of applied nitrogen. Deep placement and urease inhibitors reduced emissions by 30–70\%.  
• Citation: Xu et al. (2019). Mitigating ammonia emissions from fertilized cropping systems in China. \textit{Environmental Science \& Technology}, 53(20), 11842–11852. \url{https://doi.org/10.1021/acs.est.9b02917}

\textbf{3. Industrial Emissions (Ammonia Production \& Chemical Plants)}  
Case Study: NH$_3$ emissions from a Haber-Bosch plant in the U.S.  
• Findings: Fugitive emissions from ammonia synthesis and storage contributed ~5–10\% of total plant output. Scrubbers and catalytic converters reduced emissions by 85\%.  
• Citation: EPA (2016). National Emission Inventory (NEI) – Ammonia Emissions from Chemical Manufacturing. U.S. Environmental Protection Agency.

\textbf{4. Transportation (Vehicular NH$_3$ from Catalytic Converters)}  
Case Study: NH$_3$ emissions from gasoline vehicles in Europe  
• Findings: Three-way catalytic converters in gasoline cars emitted 10–100 mg NH$_3$/km, with higher emissions under rich fuel-air mixtures.  
• Citation: Suarez-Bertoa et al. (2015). Ammonia exhaust emissions from spark-ignition vehicles. \textit{Atmospheric Environment}, 103, 20–27. \url{https://doi.org/10.1016/j.atmosenv.2014.12.027}

\textbf{5. Waste Management (Landfills \& Wastewater Treatment)}  
Case Study: NH$_3$ release from municipal landfills in India  
• Findings: Decomposing organic waste generated 0.5–2.0 kg NH$_3$/ton of waste, with higher emissions in tropical climates. Biofilters reduced NH$_3$ by 60–80\%.  
• Citation: Nagarajan et al. (2013). Ammonia emissions from landfill sites in India. \textit{Waste Management}, 33(4), 1013–1018. \url{https://doi.org/10.1016/j.wasman.2012.12.016}

\textbf{6. Biomass Burning (Agricultural Residue \& Wildfires)}  
Case Study: NH$_3$ emissions from crop residue burning in the U.S. Midwest  
• Findings: Post-harvest burning of corn stubble released ~2.5 kg NH$_3$ per hectare, contributing to regional PM$_{2.5}$ formation.  
• Citation: Wiedinmyer et al. (2014). Estimates of NH$_3$ from biomass burning in the U.S. \textit{Environmental Science \& Technology}, 48(2), 907–915. \url{https://doi.org/10.1021/es4031587}

\subsection*{Q4: What techniques are best suited for handling missing or biased data in large-scale environmental monitoring networks, particularly in low- and middle-income regions? Provide me answers with citation}

\subsubsection*{Anwer Poruced by Emission GPT}

Handling missing or biased data in large-scale environmental monitoring networks, especially in low- and middle-income regions, requires robust techniques that can account for the limitations of data collection infrastructure. Below are some of the best-suited techniques for addressing these challenges:

\textbf{1. Statistical Imputation Methods}  
Statistical methods such as mean imputation, regression imputation, and multiple imputation are widely used to handle missing data. These methods estimate missing values based on observed data patterns. For example:
\begin{itemize}
  \item \textbf{Mean/Median Imputation}: Replaces missing values with the mean or median of the available data. While simple, this method may introduce bias if the data is not missing at random.
  \item \textbf{Regression Imputation}: Uses a regression model to predict missing values based on other variables. This approach preserves relationships between variables but can lead to overfitting if not carefully applied.
\end{itemize}
\textit{Imputed Value = }$f$(Observed Data), where $f$ represents a statistical function derived from the dataset\cite{ref1}.

\textbf{2. Machine Learning-Based Approaches}  
Machine learning models, particularly those designed for handling incomplete datasets, offer advanced solutions for imputing missing values. Techniques include:
\begin{itemize}
  \item \textbf{K-Nearest Neighbors (KNN) Imputation}: Identifies similar observations (neighbors) and uses their values to fill in missing data.
  \item \textbf{Matrix Factorization}: Decomposes the data matrix into lower-dimensional components, allowing for the estimation of missing entries.
  \item \textbf{Autoencoders}: Neural network-based models trained to reconstruct input data, which can also infer missing values during the reconstruction process.
\end{itemize}
Studies have shown that machine learning algorithms like Random Forests and Gradient Boosting Machines (GBMs) can effectively handle missing data in emission inventories by leveraging patterns in existing datasets\cite{ref2}.

\textbf{3. Remote Sensing and Satellite Data Integration}  
In regions where ground-based monitoring stations are sparse, remote sensing provides an alternative source of data. Satellite imagery can be used to estimate pollutant concentrations and fill gaps in spatial coverage. For example:
\begin{itemize}
  \item \textbf{MODIS (Moderate Resolution Imaging Spectroradiometer)}: Provides high-resolution data on aerosols, PM$_{2.5}$, and other pollutants, which can supplement ground-based measurements.
  \item \textbf{Sentinel Satellites}: Offer continuous monitoring of land use, vegetation cover, and atmospheric composition, aiding in the development of comprehensive emission inventories.
\end{itemize}
These techniques help overcome biases introduced by unevenly distributed monitoring stations\cite{ref3}.

\textbf{4. Bayesian Hierarchical Models}  
Bayesian hierarchical models are particularly useful for dealing with uncertainty and bias in environmental data. They incorporate prior knowledge about the system being studied and allow for probabilistic inference of missing or uncertain values. This approach is well-suited for complex systems with multiple interacting factors, such as air quality networks.
\begin{equation*}
  P(\text{Missing Data}|\text{Observed Data}) \propto P(\text{Observed Data}|\text{Missing Data}) \cdot P(\text{Missing Data})
\end{equation*}
This equation represents the posterior probability of missing data given observed data, calculated using Bayes' theorem\cite{ref4}.

\textbf{5. Crowdsourcing and Citizen Science}  
Crowdsourced data from mobile applications and citizen science initiatives can complement traditional monitoring efforts. By engaging local communities, these programs collect real-time data on air quality, weather conditions, and other environmental parameters. Although crowdsourced data may contain noise, advanced filtering and validation techniques can mitigate its impact\cite{ref5}.

\textbf{6. Data Fusion Techniques}  
Data fusion combines information from multiple sources to produce more accurate and complete datasets. For example, integrating satellite data with ground-based measurements and meteorological models can enhance the spatial and temporal resolution of emission estimates. Techniques such as Kalman filtering and ensemble averaging are commonly employed in this context\cite{ref6}.

\textbf{Conclusion}  
The combination of statistical imputation, machine learning, remote sensing, Bayesian modeling, crowdsourcing, and data fusion offers a comprehensive toolkit for addressing missing or biased data in large-scale environmental monitoring networks. These techniques are especially valuable in low- and middle-income regions, where resource constraints often limit the availability of high-quality data. By adopting these approaches, researchers and policymakers can improve the accuracy and reliability of environmental assessments, ultimately supporting better decision-making for sustainable development.

\subsubsection*{Anwer Poruced by GPT-4o}

Handling missing or biased data in environmental monitoring networks—especially in low- and middle-income countries (LMICs)—requires a combination of statistical imputation, data fusion, and bias correction techniques, often supported by remote sensing and machine learning. Below is a summary of effective methods, with relevant literature citations:

\textbf{1. Spatiotemporal Interpolation and Imputation}  
These methods are used to estimate missing values in time series and across spatial domains.
\begin{itemize}
  \item \textbf{Kriging and Co-Kriging}: Geostatistical interpolation that accounts for spatial correlation among monitoring stations. \\
  \textit{Citation:} Hoek, G., Beelen, R., de Hoogh, K., et al. (2008). A review of land-use regression models to assess spatial variation of outdoor air pollution. \textit{Atmospheric Environment}, 42(33), 7561–7578. \url{https://doi.org/10.1016/j.atmosenv.2008.05.057}
  \item \textbf{Spatiotemporal Bayesian Hierarchical Models}: Handle missing data by borrowing strength across space and time. \\
  \textit{Citation:} Berrocal, V. J., Gelfand, A. E., \& Holland, D. M. (2010). A spatio-temporal downscaler for outputs from numerical models. \textit{Journal of Agricultural, Biological, and Environmental Statistics}, 15, 176–197. \url{https://doi.org/10.1007/s13253-009-0004-z}
\end{itemize}

\textbf{2. Machine Learning-Based Imputation}  
\begin{itemize}
  \item \textbf{Random Forests and Gradient Boosting for missing value imputation} \\
  \textit{Citation:} Chen, T., \& Guestrin, C. (2016). XGBoost: A scalable tree boosting system. \textit{KDD}. \url{https://doi.org/10.1145/2939672.2939785}
  \item \textbf{Deep Learning (e.g., Autoencoders, RNNs)}: Used for imputing time series air quality data with temporal dependencies. \\
  \textit{Citation:} Zhang, J., Zheng, Y., \& Qi, D. (2017). Deep Spatio-Temporal Residual Networks for Citywide Crowd Flows Prediction. \textit{AAAI}.
\end{itemize}

\textbf{3. Data Fusion with Remote Sensing and Low-Cost Sensors}  
\begin{itemize}
  \item \textbf{Combining satellite-derived measurements (e.g., MODIS AOD) with sparse ground data} \\
  \textit{Citation:} van Donkelaar, A., Martin, R. V., Brauer, M., et al. (2016). Global estimates of fine particulate matter using a combined geophysical-statistical method with information from satellites, models, and monitors. \textit{Environmental Science \& Technology}, 50(7), 3762–3772. \url{https://doi.org/10.1021/acs.est.5b05833}
  \item \textbf{Low-cost sensor calibration using co-location and machine learning} \\
  \textit{Citation:} Malings, C., Tanzer, R., Hauryliuk, A., et al. (2019). Development of a general calibration model and long-term performance evaluation of low-cost sensors for air pollutant gas monitoring. \textit{Atmospheric Measurement Techniques}, 12, 903–920. \url{https://doi.org/10.5194/amt-12-903-2019}
\end{itemize}

\textbf{4. Bias Correction and Harmonization}  
\begin{itemize}
  \item \textbf{Instrument Bias Correction Models}: Empirical correction using linear regression or advanced techniques with co-located reference instruments. \\
  \textit{Citation:} Castell, N., Dauge, F. R., Schneider, P., et al. (2017). Can commercial low-cost sensor platforms contribute to air quality monitoring and exposure estimates? \textit{Environment International}, 99, 293–302. \url{https://doi.org/10.1016/j.envint.2016.12.007}
  \item \textbf{Ensemble Learning}: Aggregates predictions from multiple models to reduce bias.
\end{itemize}

\textbf{5. Community Monitoring and Crowdsourced Validation}  
\begin{itemize}
  \item \textbf{Citizen science and mobile sensing to fill data gaps} \\
  \textit{Citation:} Schneider, P., Castell, N., Lahoz, W. A., et al. (2017). Mapping urban air quality using mobile sensor networks: Can coverage be improved by augmenting fixed-site networks with mobile measurements? \textit{Environmental Science \& Technology}, 51(12), 6581–6589. \url{https://doi.org/10.1021/acs.est.6b04353}
\end{itemize}

\textbf{Conclusion}  
In LMICs where data gaps and sensor limitations are common, a multi-modal approach—combining interpolation, machine learning, satellite data fusion, and low-cost sensor calibration—offers a robust framework for improving environmental monitoring reliability and accuracy.

\subsubsection*{Anwer Poruced by DeepSeek R1}

Handling missing or biased data in large-scale environmental monitoring networks, particularly in low- and middle-income countries (LMICs), requires a combination of statistical, computational, and participatory approaches. Below are some of the most effective techniques, along with relevant citations:

\textbf{1. Imputation Techniques for Missing Data}  
\begin{itemize}
  \item \textbf{Multiple Imputation (MI)}: Generates multiple plausible values for missing data points, accounting for uncertainty. \\
  \textit{Citation:} Rubin, D. B. (1987). \textit{Multiple Imputation for Nonresponse in Surveys}. Wiley.
  \item \textbf{K-Nearest Neighbors (KNN)}: Uses similarity measures to estimate missing values based on nearby stations. \\
  \textit{Citation:} Troyanskaya, O., et al. (2001). \textit{Bioinformatics}, 17(6), 520–525.
  \item \textbf{Expectation-Maximization (EM)}: Iteratively estimates missing values using maximum likelihood. \\
  \textit{Citation:} Dempster, A. P., et al. (1977). \textit{Journal of the Royal Statistical Society}, 39(1), 1–38.
  \item \textbf{Machine Learning (Random Forest, XGBoost)}: Effective for nonlinear relationships. \\
  \textit{Citation:} Jäger, S., et al. (2021). \textit{Environmental Modelling \& Software}, 135, 104926.
\end{itemize}

\textbf{2. Bias Correction Methods}  
\begin{itemize}
  \item \textbf{Quantile Mapping}: Adjusts biased data distributions to match reference datasets. \\
  \textit{Citation:} Themeßl, M. J., et al. (2012). \textit{Climate Dynamics}, 38(1–2), 155–173.
  \item \textbf{Bayesian Hierarchical Models}: Integrates sparse observations with satellite or reanalysis data. \\
  \textit{Citation:} Banerjee, S., et al. (2014). \textit{Hierarchical Modeling and Analysis for Spatial Data}. CRC Press.
  \item \textbf{Sensor Calibration \& Crowdsourcing}: Uses low-cost sensors alongside reference stations. \\
  \textit{Citation:} Snyder, E. G., et al. (2019). \textit{Atmospheric Environment}, 203, 1–11.
\end{itemize}

\textbf{3. Hybrid Monitoring \& Citizen Science}  
\begin{itemize}
  \item \textbf{Satellite Data Fusion}: Combines ground measurements with remote sensing. \\
  \textit{Citation:} Gorelick, N., et al. (2017). \textit{Remote Sensing of Environment}, 202, 18–27.
  \item \textbf{Mobile Sensors \& IoT}: Low-cost sensors deployed on vehicles or drones. \\
  \textit{Citation:} Kumar, P., et al
\end{itemize}

\subsection*{Q5: Can you share case studies illustrating how dust sources have been identified and quantified through source apportionment methods}

\subsubsection*{Answer Produced by Emission GPT}

Dust sources are a significant contributor to particulate matter (PM) pollution, and source apportionment studies have been conducted in various regions to identify and quantify their contributions. Below are some cases of source apportionment from dust sources based on the provided references.

\textbf{Case 1: Soil Dust and Road Dust in Xi’an, China}\\
In a study conducted in Xi’an, China, the PMF (Positive Matrix Factorization) model was used to analyze the sources of fine particulate matter (PM$_{2.5}$). The results indicated that soil dust and road dust were major contributors to PM$_{2.5}$. Specifically, soil dust contributed significantly during dry seasons when wind erosion is more prevalent, while road dust was prominent in urban areas with high traffic volumes~\cite{ref1}.
\begin{equation}
    c_{ijk} = \sum_{p=1}^{P} g_{ipk} f_{pjk} + e_{ijk}
\end{equation}
Where $c_{ijk}$ represents the concentration of the $j$-th component in the $i$-th sample for the $k$-th particle size, $g_{ipk}$ is the contribution of the $p$-th source to the $i$-th sample for the $k$-th particle size, $f_{pjk}$ is the fraction of the $j$-th component in the $p$-th source for the $k$-th particle size, and $e_{ijk}$ is the residual.

\textbf{Case 2: Construction Dust in Beijing, China}\\
A study in Beijing utilized the CMB (Chemical Mass Balance) model to apportion sources of PM$_{2.5}$. Construction dust was identified as one of the primary contributors, especially during periods of active construction. The study found that construction dust accounted for approximately 16\% of the total PM$_{2.5}$ mass, highlighting the need for stricter regulations on construction activities to mitigate dust emissions~\cite{ref2}.

\textbf{Case 3: Industrial Dust in Shandong Province, China}\\
In Shandong Province, a comprehensive study analyzed the sources of PM$_{2.5}$ using multiple models, including PMF and CMB. Industrial dust was found to be a significant contributor, particularly in industrialized cities. The study categorized industrial dust into specific sectors such as metallurgical dust, steel dust, and cement dust, each contributing differently based on local industrial activities~\cite{ref3}.
\begin{equation}
    E_i = A_i - D_i
\end{equation}
Where PM$_{2.5,0}$ represents the contribution of non-dust sources, PM$_{2.5,1}$ is the total contribution, and PM$_{2.5,2}$ is the contribution from dust sources.

\textbf{Case 4: Agricultural Dust in Rural Areas}\\
Agricultural activities can also generate significant amounts of dust, especially in rural areas. A study in a rural region of China used receptor models to identify agricultural dust as a major source of coarse particles (PM$_{2.5-3}$). The study highlighted the importance of managing agricultural practices to reduce dust emissions, such as implementing conservation tillage and controlling livestock grazing~\cite{ref4}.

\textbf{Conclusion}\\
These case studies demonstrate the diversity of dust sources and their contributions to particulate matter pollution. The methods used, such as PMF, CMB, and receptor models, provide valuable insights into the specific contributions of different dust sources, aiding in the development of targeted mitigation strategies.

\begin{enumerate}
    \item Xu, Y., Liu, L., Zhang, J., et al. (2022). Analysis of Winter Heavy Pollution Events in Xuzhou Based on Multi-Source Integrated Observation. \textit{Environmental Monitoring China}, 38(6), 1–10. \textit{(In Chinese)}.
    \item Zhang, R., \& Wang, Y. (2021). \textit{Atmospheric Pollution Control Engineering} (4th ed.). Beijing: Chemical Industry Press. \textit{(In Chinese)}.
    \item Liu, H., Sun, Z., \& Wang, J. (2011). A Review of PM$_{2.5}$ Source Apportionment Studies in Shandong Province. \textit{Environmental Science and Management}, 36(3), 45–49. \textit{(In Chinese)}.
    \item Zhang, W., \& Chen, M. (2010). Source Identification and Characteristics of Coarse Particulate Matter from Agricultural Activities. \textit{Environmental Research and Monitoring}, 23(4), 57–61. \textit{(In Chinese)}.
\end{enumerate}

\subsubsection*{Answer Produced by GPT-4o}

Dust source apportionment has been widely studied using models such as Chemical Mass Balance (CMB), Positive Matrix Factorization (PMF), and receptor-based methods. These are often coupled with tracer elements (e.g., Al, Si, Ca, Fe, Ti), meteorological data, and satellite observations to distinguish between natural and anthropogenic dust sources. Below are representative case studies from different regions:

\textbf{Case 1: Urban Dust vs. Regional Dust Transport in Beijing, China}\\
\textbf{Method:} PMF + Backward trajectory analysis\\
\textbf{Findings:} Two key sources identified: (1) local urban dust resuspension, and (2) long-range dust transport from Inner Mongolia and the Gobi Desert.\\
\textbf{Tracers:} Al, Si, Ca, and Ti associated with crustal sources.\\
\textit{Reference:} Zhang, R., et al. (2013). Source apportionment of PM$_{2.5}$ in Beijing using PMF and chemical tracer analysis. \textit{Science of the Total Environment}, 456--457, 1--8. https://doi.org/10.1016/j.scitotenv.2013.03.009

\textbf{Case 2: Natural vs. Anthropogenic Dust in Cairo, Egypt}\\
\textbf{Method:} CMB + X-ray fluorescence (XRF) elemental analysis\\
\textbf{Findings:} Dust contributed ~40\% of PM$_{10}$; both Saharan dust storms and local construction were significant sources.\\
\textbf{Tracers:} High levels of Si, Al, and Fe for Saharan dust; Ca for construction activity.\\
\textit{Reference:} Elserougy, S., et al. (2014). Chemical characterization and source apportionment of ambient PM$_{10}$ in Cairo, Egypt. \textit{Atmospheric Pollution Research}, 5(4), 610--618. https://doi.org/10.5094/APR.2014.069

\textbf{Case 3: Sahara Dust vs. Biomass Burning in Niger, West Africa}\\
\textbf{Method:} Gravimetric filter analysis + MODIS AOD + PMF\\
\textbf{Findings:} Sahara dust dominated PM$_{10}$ in the dry season (~70\%); biomass burning increased during the wet season.\\
\textbf{Application:} Relevant for assessing climate and health effects.\\
\textit{Reference:} Marticorena, B., et al. (2010). Exploration of the relationship between mineral dust content and size distribution with source types in the Sahel. \textit{Journal of Geophysical Research}, 115(D7). https://doi.org/10.1029/2009JD012516

\textbf{Case 4: Road Dust vs. Natural Soil in Southwestern United States}\\
\textbf{Method:} PMF + elemental ratios + local soil reference samples\\
\textbf{Findings:} Differentiated vehicular road dust (enriched in Zn, Cu, Pb) from natural dust (rich in Si, Al, Ca).\\
\textbf{Policy Relevance:} Results informed decisions on paving unpaved roads in tribal areas.\\
\textit{Reference:} Hand, J. L., et al. (2017). Source apportionment of fine soil mass in PM$_{2.5}$ across the United States. \textit{Atmospheric Environment}, 148, 132--145. https://doi.org/10.1016/j.atmosenv.2016.10.044

\textbf{Case 5: Construction Dust and Regional Intrusion in Delhi, India}\\
\textbf{Method:} PMF + MODIS AOD + local emission inventory\\
\textbf{Findings:} Local construction was a dominant source; long-range dust from the Thar Desert also contributed, especially in summer.\\
\textit{Reference:} Guttikunda, S. K., \& Gurjar, B. R. (2012). Role of meteorology in seasonality of air pollution in megacity Delhi, India. \textit{Environmental Monitoring and Assessment}, 184(5), 3199--3211. https://doi.org/10.1007/s10661-011-2158-1

\textbf{Case 6: Saharan Dust in the Mediterranean Basin}\\
\textbf{Method:} Chemical and mineralogical analysis + PMF + air mass trajectories\\
\textbf{Findings:} Episodic Saharan dust transport increased PM$_{10}$ across southern Europe, including Italy, Spain, and Greece.\\
\textbf{Tracers:} Enrichment factors (EFs), isotopic signatures (e.g., Sr/Nd ratios).\\
\textit{Reference:} Querol, X., et al. (2009). African dust contributions to mean ambient PM$_{10}$ mass-levels across the Mediterranean Basin. \textit{Atmospheric Environment}, 43(28), 4266--4277. https://doi.org/10.1016/j.atmosenv.2009.06.013

\subsubsection*{Answer Produced by DeepSeek R1}

Certainly! Dust source identification and quantification rely on \textbf{source apportionment methods}, which combine field measurements, remote sensing, and modeling.

Below are key case studies demonstrating these approaches, with citations for further reading.

\bigskip

\textbf{1. Saharan Dust Outbreaks (North Africa)} \\
\textbf{Method:} Satellite Remote Sensing + HYSPLIT Modeling \\
\textbf{Study:} Identification of Saharan dust sources using \textbf{MODIS Deep Blue aerosol products} and \textbf{HYSPLIT backward trajectories}. \\
\textbf{Findings:} Major dust sources were linked to \textbf{dried lake beds (chotts)}, \textbf{alluvial deposits}, and \textbf{ergs} (sand seas). \\
\textbf{Quantification:} Dust emission fluxes were estimated using \textbf{DREAM (Dust REgional Atmospheric Model)}. \\
\textbf{Citation:} \\
\hspace*{1em} Prospero et al. (2002) – \textit{Journal of Geophysical Research}, DOI: \href{https://doi.org/10.1029/2001JD000963}{10.1029/2001JD000963}. \\
\hspace*{1em} Schepanski et al. (2007) – \textit{Journal of Geophysical Research}, DOI: \href{https://doi.org/10.1029/2006JD007190}{10.1029/2006JD007190}.

\bigskip

\textbf{2. Gobi Desert Dust (East Asia)} \\
\textbf{Method:} Chemical Mass Balance (CMB) + Positive Matrix Factorization (PMF) \\
\textbf{Study:} Dust from the Gobi Desert was analyzed in \textbf{Beijing} using \textbf{PMF} on PM\textsubscript{10} data. \\
\textbf{Findings:} \\
\hspace*{1em} \textasciitilde 40\% of dust events originated from \textbf{southern Mongolia}. \\
\hspace*{1em} Local sources (construction, soil resuspension) contributed \textasciitilde 30\%. \\
\textbf{Quantification:} \textbf{Dust deposition rates} were estimated using \textbf{WRF-Chem}. \\
\textbf{Citation:} \\
\hspace*{1em} Wang et al. (2015) – \textit{Atmospheric Chemistry and Physics}, DOI: \href{https://doi.org/10.5194/acp-15-1999-2015}{10.5194/acp-15-1999-2015}. \\
\hspace*{1em} Zhang et al. (2018) – \textit{Science of the Total Environment}, DOI: \href{https://doi.org/10.1016/j.scitotenv.2017.09.108}{10.1016/j.scitotenv.2017.09.108}.

\bigskip

\textbf{3. Middle Eastern Dust (Iraq \& Iran)} \\
\textbf{Method:} Dust Storm Index (DSI) + Geochemical Fingerprinting \\
\textbf{Study:} Dust sources in the \textbf{Tigris-Euphrates Basin} were mapped using \textbf{DSI (MODIS + AERONET)} and \textbf{Sr-Nd isotopic ratios}. \\
\textbf{Findings:} \\
\hspace*{1em} \textbf{Dust hotspots} were traced to \textbf{dried marshlands (Ahwar)} and \textbf{agricultural fallow lands}. \\
\hspace*{1em} \textbf{Iranian dust plumes} were enriched in \textbf{gypsum}, distinguishing them from Iraqi sources. \\
\textbf{Citation:} \\
\hspace*{1em} Al-Dousari et al. (2013) – \textit{Arabian Journal of Geosciences}, DOI: \href{https://doi.org/10.1007/s12517-012-0701-8}{10.1007/s12517-012-0701-8}. \\
\hspace*{1em} Rashki et al. (2021) – \textit{Science of the Total Environment}, DOI: \href{https://doi.org/10.1016/j.scitotenv.2020.143009}{10.1016/j.scitotenv.2020.143009}.

\bigskip

\textbf{4. Australian Dust (Lake Eyre Basin)} \\
\textbf{Method:} Dust Deposition Sampling + Trajectory Analysis \\
\textbf{Study:} Dust from \textbf{Lake Eyre Basin} was tracked to \textbf{New Zealand glaciers} using \textbf{mineralogical tracers (quartz, clay ratios)}. \\
\textbf{Findings:} \\
\hspace*{1em} \textbf{Ephemeral lakes} contributed >50\% of long-range dust. \\
\hspace*{1em} \textbf{Dust emission peaks} correlated with \textbf{drought cycles (ENSO-driven)}. \\
\textbf{Citation:} \\
\hspace*{1em} Marx et al. (2018) – \textit{Geophysical Research Letters}, DOI: \href{https://doi.org/10.1029/2018GL079298}{10.1029/2018GL079298}.

\bigskip

\textbf{5. Thar Desert Dust (India \& Pakistan)} \\
\textbf{Method:} PCA (Principal Component Analysis) + HYSPLIT \\
\textbf{Study:} Dust storms in \textbf{Rajasthan} were analyzed using \textbf{PCA on PM\textsubscript{2.5} composition}. \\
\textbf{Findings:} \\
\hspace*{1em} \textbf{Local crustal sources} dominated (\textasciitilde 60\%), while \textbf{anthropogenic dust (coal, industry)} contributed \textasciitilde 25\%. \\
\hspace*{1em} \textbf{Long-range transport} from the \textbf{Arabian Peninsula} was detected via \textbf{back-trajectory clustering}. \\
\textbf{Citation:} \\
\hspace*{1em} Kumar et al. (2020) – \textit{Atmospheric Environment}, DOI: \href{https://doi.org/10.1016/j.atmosenv.2020.117410}{10.1016/j.atmosenv.2020.117410}.

\bigskip

\textbf{Key Takeaways} \\
\begin{itemize}
    \item \textbf{Remote Sensing (MODIS, AERONET)} $\rightarrow$ Identifies dust plumes \& hotspots.
    \item \textbf{Chemical Fingerprinting (PMF, CMB, isotopes)} $\rightarrow$ Distinguishes natural vs. anthropogenic sources.
    \item \textbf{Trajectory Models (HYSPLIT, WRF-Chem)} $\rightarrow$ Tracks dust transport pathways.
    \item \textbf{Geostatistical Methods (PCA, DSI)} $\rightarrow$ Quantifies source contributions.
\end{itemize}

\newpage

\bibliographystyle{unsrt}  
\bibliography{references}  

%
%
%
%
\end{document}